\documentclass[conference]{IEEEtran}
\IEEEoverridecommandlockouts
\usepackage{cite}
\usepackage{amsmath,amssymb,amsfonts}
\usepackage{algorithmic}
\usepackage{graphicx}
\usepackage{textcomp}
\usepackage{microtype}
\usepackage{graphicx}
\usepackage{subfigure}
\usepackage{tablefootnote}
\usepackage{multirow}
\usepackage{array}
\usepackage{enumitem}
\usepackage{booktabs} 
\usepackage[table]{xcolor}
\usepackage{colortbl}
\usepackage{balance}

\def\BibTeX{{\rm B\kern-.05em{\sc i\kern-.025em b}\kern-.08em
    T\kern-.1667em\lower.7ex\hbox{E}\kern-.125emX}}
\makeatletter
    \newcommand{\linebreakand}{%
      \end{@IEEEauthorhalign}
      \hfill\mbox{}\par
      \mbox{}\hfill\begin{@IEEEauthorhalign}
    }
\makeatother
\begin{document}

\title{Learning Multi-Pattern Normalities in the Frequency Domain for Efficient Time Series Anomaly Detection
\thanks{$^\dagger$Corresponding author.}
}

\author{\IEEEauthorblockN{Feiyi Chen$^{1}$, Yingying Zhang$^{2}$, Zhen Qin$^1$, Lunting Fan$^2$, Renhe Jiang$^3$, Yuxuan Liang$^4$, \\Qingsong Wen$^5$, Shuiguang Deng$^1$$^\dagger$}
\textit{$^1$Zhejiang University} \quad
\textit{$^2$Alibaba Group}  \quad
\textit{$^3$The University of Tokyo}  \\
\textit{$^4$The Hong Kong University of Science and Technology (Guangzhou)}\quad
\textit{$^5$Squirrel AI} \\
\textit{$^1$\{chenfeiyi,zhenqin,dengsg\}@zju.edu.cn} \\
\textit{$^2$congrong.zyy@alibaba-inc.com} \quad
\textit{$^2$lunting.fan@taobao.com}
\\
\textit{$^3$jiangrh@csis.u-tokyo.ac.jp} \quad
\textit{$^4$yuxliang@outlook.com} \quad
\textit{$^5$qingsongedu@gmail.com}
}



\maketitle

\begin{abstract} Anomaly detection significantly enhances the robustness of cloud systems. While neural network-based methods have recently demonstrated strong advantages, they encounter practical challenges in cloud environments: the contradiction between the impracticality of maintaining a unique model for each service and the limited ability to deal with diverse normal patterns by a unified model, as well as issues with handling heavy traffic in real time and short-term anomaly detection sensitivity.
    Thus, we propose MACE, a multi-normal-pattern accommodated and efficient anomaly detection method in the frequency domain for time series anomaly detection. There are three novel characteristics of it: (i) a pattern extraction mechanism excelling at handling diverse normal patterns with a unified model, which enables the model to identify anomalies by examining the correlation between the data sample and its service normal pattern, instead of solely focusing on the data sample itself; (ii) a dualistic convolution mechanism that amplifies short-term anomalies in the time domain and hinders the reconstruction of anomalies in the frequency domain, which enlarges the reconstruction error disparity between anomaly and normality and facilitates anomaly detection; (iii) leveraging the sparsity and parallelism of frequency domain to enhance model efficiency. We theoretically and experimentally prove that using a strategically selected subset of Fourier bases can not only reduce computational overhead but is also profitable to distinguish anomalies, compared to using the complete spectrum.  Moreover, extensive experiments demonstrate MACE's effectiveness in handling diverse normal patterns with a unified model and it achieves state-of-the-art performance with high efficiency.
\end{abstract}


\section{Introduction}
\label{seg:intro}
\textcolor{black}{Anomaly detection is an extensively researched issue crucial for bolstering cloud system reliability and curbing labor expenses \cite{zhang2023dbcatcher,chen2023lara}. Reconstruction-based methods, in particular, have demonstrated state-of-the-art performance in this domain \cite{su2019robust, DBLP:conf/iclr/XuWWL22, DBLP:conf/icml/ChenTCDDZ22}. Despite these advancements, several significant challenges persist, as outlined below:}
\begin{figure*}[t]
    \centering 
    \subfigure[Data visualization of services]{
    \includegraphics[width=0.27\linewidth]{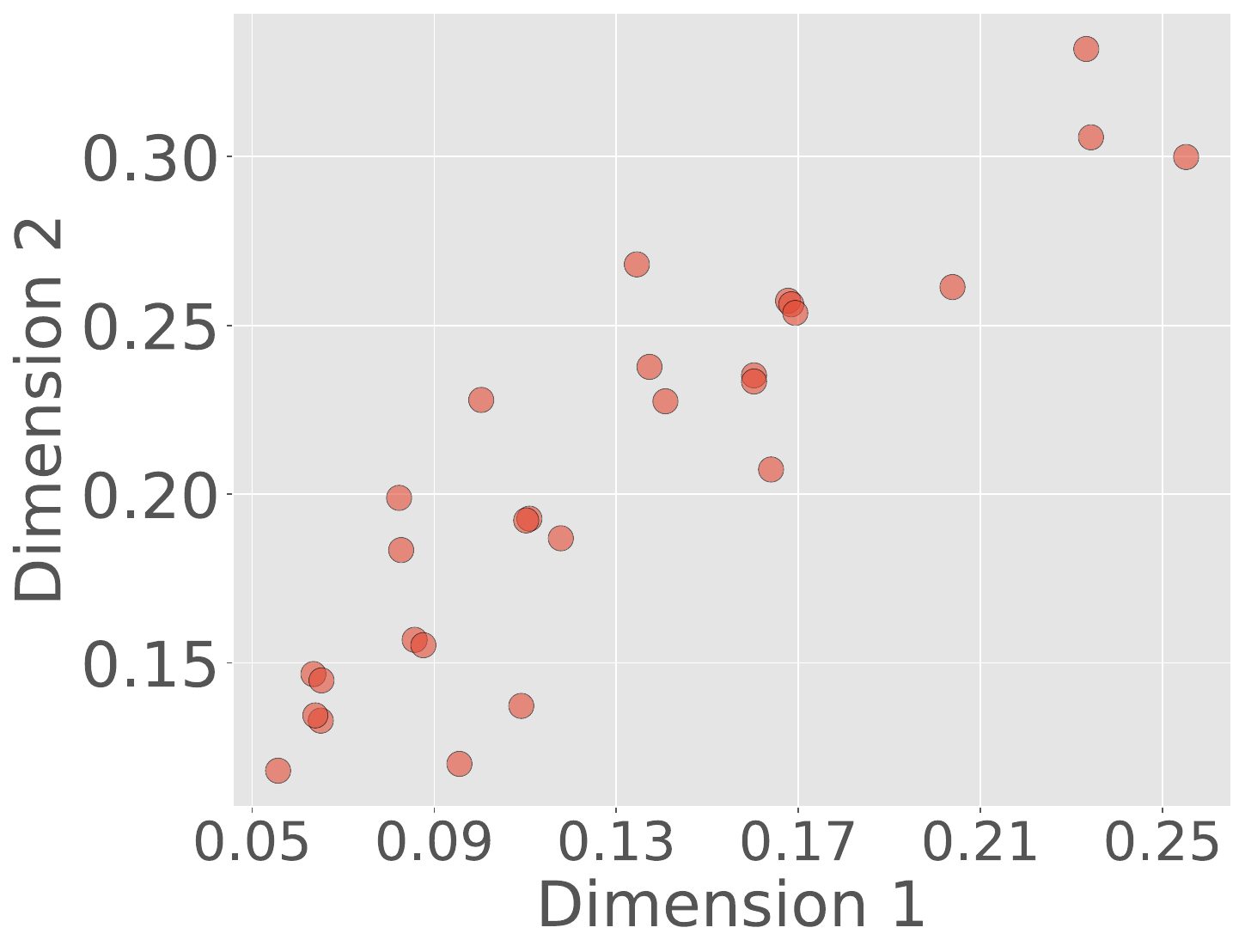}
    \label{fig:MultiPat}
    }
    \hfill
    \subfigure[The F1 score for unified model and tailored model on Server Machine Dataset]{
    \includegraphics[width=0.27\linewidth]{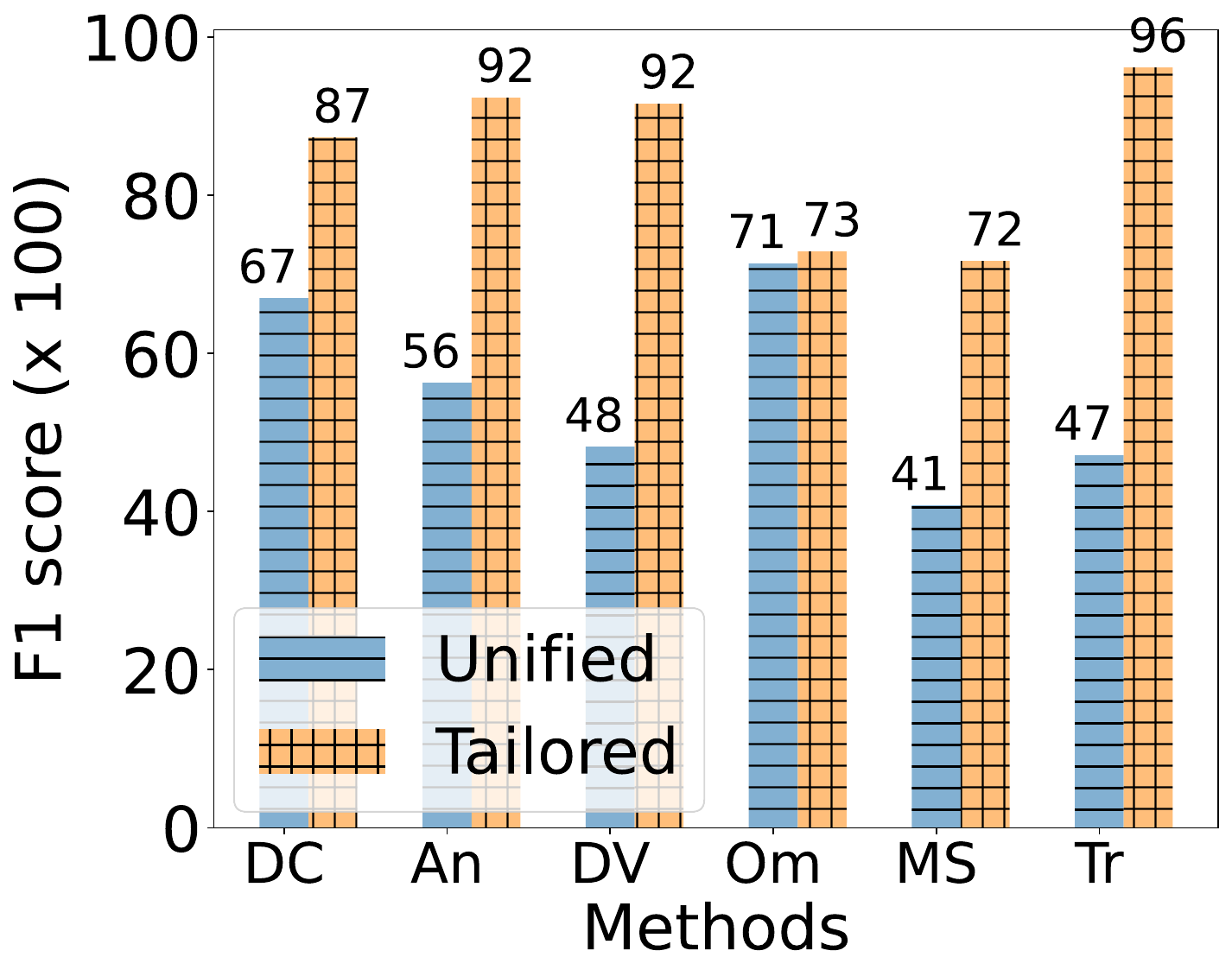}
    \label{fig:unified_tailor}
    }
    \hfill
    \subfigure[An intuitive illustration for pattern extraction mechanism.]{
    \includegraphics[width=0.34\linewidth]{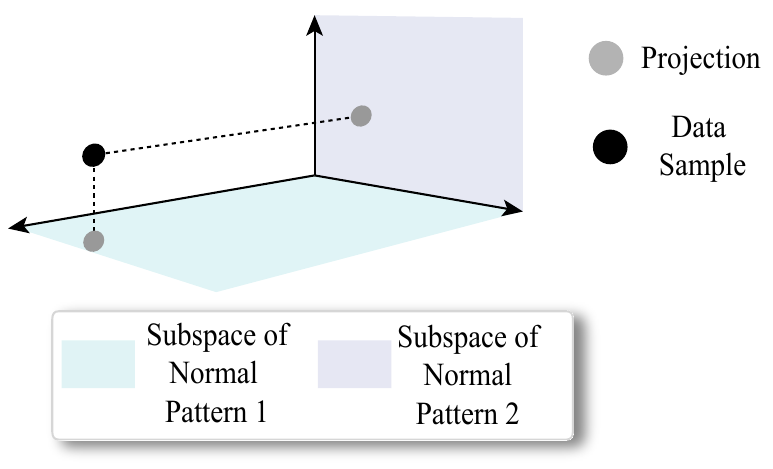}
    \label{fig:paExt}
    }
    \hfill
    \caption{(a) The normal data of each service is compressed into a two-dimensional vector, which is scattered randomly. (b) The figure shows F1 score of some SOTA methods: DCdetector \cite{DBLP:conf/kdd/YangZZW023}, AnomalyTransformer \cite{DBLP:conf/iclr/XuWWL22}, DVGCRN \cite{DBLP:conf/icml/ChenTCDDZ22}, OmniAnomaly \cite{su2019robust}, MSCRED \cite{zhang2019deep}, TranAD \cite{DBLP:journals/pvldb/TuliCJ22}. (c) A data sample is projected to a normal pattern subspace in a pattern extraction mechanism. When the data sample is closer to the normal pattern subspace, it is easier to reconstruct it from its projection with less reconstruction error. Thus, the data sample is more likely to be inferred as normality for the normal pattern 1 than normal pattern 2. anomalies.}
\end{figure*}
\begin{itemize}
\item \textbf{C1.} \emph{Limited capacity to accommodate diverse normal patterns}: In practical scenarios, cloud centers host millions of services concurrently, and each service exhibits a unique normal pattern. To illustrate this, we project the normal data of each service in the Server Machine Dataset onto a two-dimensional space and visualize their distribution in Fig.~\ref{fig:MultiPat}, where the data points are scattered randomly. It is impractical and excessively expensive to maintain a tailored model for each service\textcolor{black}{\cite{DBLP:conf/icml/LiCCWTZ23}}. However, most state-of-the-art methods train a tailored model for each cloud service \cite{su2019robust, zhang2019deep, DBLP:conf/iclr/XuWWL22}. Moreover, it is reported that many reconstruction-based methods can only effectively capture a few dominant normal patterns in the training set for each trained model \cite{park2020learning}. We further verify this by conducting an experiment comparing performances of some state-of-the-art methods when training a unified model for ten services and tailoring ten models for ten services, as shown in Fig.\ref{fig:unified_tailor}. The performance of the unified model is substantially lower than the one of the tailored models.

\item \textbf{C2.} \emph{Inefficiency in handling heavy traffic in real time}: In large cloud centers, the volume of service traffic can escalate to hundreds of thousands of requests per second. In these high-demand scenarios, numerous deep learning-based methods face challenges in efficiently handling peak traffic in real time. Furthermore, a notable complication arises from the incorporation of recurrent networks in several anomaly detection neural networks, such as VRNN \cite{chung2015recurrent}, omniAnomaly \cite{su2019robust}, and MSCRED \cite{zhang2019deep}. This inclusion hampers operator parallelization, as recurrent networks cannot be effectively parallelized across different recurrent steps. 

\item \textbf{C3.} \emph{Insensitivity to single point anomalies}: Encoder-decoder neural networks, which are used by many reconstruction-based methods, are insensitive to single point anomalies and often overlook them \cite{DBLP:journals/pvldb/TuliCJ22,schmidl2022anomaly}. 
\end{itemize}

Tackling these challenges is imperative for enhancing the efficacy and applicability of deep learning-based anomaly detection methods in practical cloud environments. Therefore, we design innovative mechanisms to address these issues, as outlined below:
\begin{itemize}
    \item \textbf{S1.} To enhance the model's ability to accommodate diverse normal patterns (C1), we propose a pattern extraction mechanism. \emph{The most challenging problem of dealing with diverse normal patterns is that an anomaly for one normal pattern could be a normality for another.} Thus, we detect anomalies according to the correlation between the data sample and its service normal pattern, instead of the data sample itself.
    The pattern extraction mechanism identifies a normal pattern subspace in the frequency domain for each service. Subsequently, it tailors a representation for each data sample from the sample's projection on its service normal pattern subspace. In this way, when the data sample is closer to its service normality subspace, it is easier to reconstruct from the representation with less reconstruction error and is more likely to be inferred as normality, as shown in Fig. \ref{fig:paExt}.
    \item \textbf{S2.} \textcolor{black}{To enhance model efficiency and parallelism (C2), we introduce a frequency-domain-based approach. Anomaly detection in the frequency domain can leverage sparsity to reduce computational overhead and enhance fine-grained parallelism by eliminating temporal dependencies. To exploit frequency domain sparsity, we propose a strategy for selecting a subset of Fourier bases for each service in the pattern extraction process. We theoretically demonstrate that using only this subset of Fourier bases not only reduces computational overhead but also improves anomaly detection performance compared to utilizing the complete spectrum. For effective anomaly detection in the frequency domain, we introduce a dualistic convolution mechanism to replace the standard convolution in the auto-encoder. This mechanism hinders the reconstruction of anomalies while keeping the reconstruction of normalities easy.}
    \item \textbf{S3.} To enhance the sensitivity to short-term anomalies (C3), we introduce the dualistic convolution mechanism to the time domain, which amplifies the anomalies, facilitating their detection while maintaining the similarity of normality to the original time series. As illustrated in Fig. \ref{fig:ampl2}, the dualistic convolution mechanism extends anomalies while maintaining the similarity of normality to the original time series. 
\end{itemize}

Accordingly, this work makes the following novel and unique contributions to the field of anomaly detection:
\begin{itemize}
    \item We propose a novel pattern extraction mechanism to deal with diverse normal patterns by facilitating the model to detect anomalies from the correlation between the data sample and its service normal pattern, instead of only the data sample.
    \item We propose a dualistic convolution mechanism. In the time domain, it amplifies the anomalies. In the frequency domain, it hinders the reconstruction of anomalies while keeping the reconstruction of normalities easy.
    \item We leverage the sparsity and parallelism of the frequency domain to improve model efficiency. It is theoretically and experimentally proved that using just a subset of Fourier bases can not only reduce computational overhead but also achieve better anomaly detection performance compared with using the complete spectrum.
\end{itemize}
Moreover, we conduct extensive experiments on four real-world datasets to demonstrate that MACE can effectively capture diverse normal patterns with a unified model and averagely improve F1 score by 8.7\%, compared with the strongest method in baselines, while achieving 4$\times$ faster.

\section{Related Work}
\label{seg:rela}
\textcolor{black}{
Anomaly detection is a crucial task that focuses on identifying outliers within time series data and has been the subject of extensive research. The existing body of work can be broadly categorized into three main groups: classical machine learning \cite{ramaswamy2000efficient,wang2018exact,gomez2011adaptive} and statistical methods \cite{boniol2021unsupervised,zhang2021cloudrca,siffer2017anomaly,subramaniam2006online}, signal processing-based methods \cite{ren2019time,thill2018online,thill2017time}, and deep learning-based methods \cite{gao2020robusttad, chen2021daemon,zhang2022tfad,li2022learning,goodge2022lunar,jin2023survey}. In the following, we provide a brief overview of each category and a review of multi-task learning since multiple normal pattern learning is highly correlated to the aim of this work.}

\subsection{Anomaly Detection}
\textcolor{black}{
\textbf{Classical methods.} 
Conventional statistical and machine learning methods, as highlighted in earlier works \cite{choffnes2010crowdsourcing, guha2016robust, pang2015lesinn}, operate without the need for extensive training data and remain unaffected by the challenge of diverse normal patterns. Furthermore, they typically incur little computational overhead. Despite these advantages, these approaches are contingent upon certain assumptions and, in real-world applications, exhibit constrained robustness \cite{ma2021jump}.
}

\textcolor{black}{
\textbf{Signal-processing-based methods.} 
Signal processing-based methods leverage the advantages of fine-grained parallelism and the sparsity inherent in the frequency domain. Despite these advantages, they encounter difficulties in simultaneously capturing both global and subtle features while maintaining manageable computational overhead. For instance, the Fourier transform \cite{zhao2019automatic} excels at capturing global information but struggles with subtle local features. In contrast, context-aware Discrete Fourier Transform (DFT) and context-aware Inverse DFT (IDFT) select Fourier bases based on a given normal pattern and integrate with dualistic convolution mechanism, enhancing the capacity to extract subtle features. The theoretical evidence supports that this approach widens the reconstruction error gap between anomalies and normal patterns.
JumpStarter \cite{ma2021jump}, a recent state-of-the-art method in this category, suffers from significant inference time overhead and struggles to handle heavy traffic loads in real time.}

\textcolor{black}{
\textbf{Deep learning-based methods.} 
Deep learning-based methods prove to be particularly effective for variable time series \cite{DBLP:journals/pvldb/TuliCJ22}. These methods can be broadly categorized into prediction-based approaches \cite{hundman2018detecting, DBLP:conf/iclr/ZongSMCLCC18}, reconstruction-based methods \cite{DBLP:conf/iclr/XuWWL22, DBLP:conf/icml/ChenTCDDZ22}, and classifier-based methods \cite{DBLP:conf/iclr/GrathwohlWJD0S20, DBLP:conf/icml/RuffGDSVBMK18, shen2020timeseries}. Prediction-based methods, such as LSTM-NDT \cite{hundman2018detecting} and DAGMM \cite{DBLP:conf/iclr/ZongSMCLCC18}, incorporate recurrent networks, which are non-parallelizable and inefficient. Similarly, reconstruction-based methods like Donut \cite{xu2018unsupervised} and OmniAnomaly \cite{su2019robust} also rely on recurrent neural networks. Recent advancements, exemplified by USAD \cite{audibert2020usad} and GDN \cite{deng2021graph}, replace recurrent neural networks with attention-based architectures to expedite the training process. However, these methods face challenges in effectively capturing long-term dependencies due to the removal of recurrent networks and the use of small input windows \cite{DBLP:journals/pvldb/TuliCJ22}.
In contrast to these approaches, anomaly detection in the frequency domain eliminates temporal dependencies without sacrificing global information. Consequently, there is no need for recurrent neural networks, yet the model can still effectively capture long-term features.
The most recent works leverage the power of transformers, exemplified by AnomalyTransformer \cite{DBLP:conf/iclr/XuWWL22} and TranAD \cite{DBLP:journals/pvldb/TuliCJ22}, enabling fine-grained parallelism. However, these methods encounter challenges in handling diverse normal patterns with a unified model.}

\subsection{Multi-task Learning}
\textcolor{black}{
Multitask Learning (MTL) is a machine learning paradigm in which a model is trained to concurrently address multiple interrelated tasks. Unlike the conventional approach of training separate models for individual tasks, MTL facilitates the sharing of specific model parameters. This sharing mechanism allows the model to glean common representations across the spectrum of tasks it is designed to handle. Within the realm of MTL research, two principal categories emerge: hard-sharing methods and soft-sharing methods.
Hard-sharing methods involve the sharing of common low-level hidden layers among the tasks. In contrast, soft-sharing methods adopt a more nuanced strategy. They foster the exchange of general knowledge among multiple models by incorporating regularization techniques into neural network parameters \cite{DBLP:conf/iclr/YangH17} or establishing connections across networks \cite{liu2020adaptive,ma2019snr}.
However, both soft-sharing and hard-sharing methods necessitate the maintenance of task-specific neural network layers. This maintenance, despite the advancements in MTL, remains computationally expensive, particularly when dealing with a multitude of diverse services and tasks in real-world applications. Finding more efficient ways to share knowledge across tasks without compromising performance is an ongoing challenge.} 


\section{Preliminary}
A normal multivariate time series is denoted by $X_N \in \mathbb{R}^{T\times m}$, where $T$ denotes the sliding window length and $m$ denotes the number of feature dimensions. Similarly, an anomalous multivariate time series is denoted by $X_A \in \mathbb{R}^{T\times m}$. \textcolor{black}{A normal pattern is a distribution that determines the possibility $p(X_N[t]|X_N[1:t])$, where $X_N[t]$ denotes the $t^{th}$ element of $X_N$ and $X_N[1:t]$ denotes the subsequence of $X_N$ from the first slot to the $(t-1)^{th}$ slot.}
The reconstruction-based method compresses $X_N$ and $X_A$ separately and reconstructs them. The reconstructed normal time series is denoted by $\tilde{X}_N$, while the reconstructed anomalous one is denoted by $\tilde{X}_A$. The objective of reconstruction-based methods is shown in Eq.~\ref{Eq:obj}, which enlarges the gap between the reconstruction error of normalities and anomalies. Subsequently, it determines a threshold by some strategies, such as POT \cite{siffer2017anomaly}. When the reconstruction error exceeds the threshold, the input data is inferred as an anomaly.
\begin{equation}
    \max. |\tilde{X}_A-X_A| - |\tilde{X}_N-X_N|
    \label{Eq:obj}
\end{equation}

Additionally, other symbols utilized in this paper are outlined in Table \ref{Tab:symb}.
\begin{table}[]
    \centering
    \renewcommand\arraystretch{1}
    \caption{\label{Tab:symb}The definition of symbols.}
    \begin{tabular}{c|c}
        \hline
        \textbf{Symbol}           & \textbf{Definition}                                                \\ \hline
        $\gamma$                  & The power in dualistic convolution                                 \\
        $\sigma$                  & The scaling factor in dualistic convolution                        \\
        $s$                       & The stride length of convolution                                   \\
        $\omega_i$                & The frequency of $i^{th}$ strongest signal of normal pattern       \\
        $\mathcal{A}_N(\omega_i)$ & The amplitude of normality spectrum corresponding to $\omega_i$    \\
        $\mathcal{A}_A(\omega_i)$ & The amplitude of anomaly spectrum corresponding to $\omega_i$      \\
        $q_N(\omega_i)$           & The normalized value of $\mathcal{A}_N(\omega_i)$                  \\
        $q_A(\omega_i)$           & The normalized value of $\mathcal{A}_A(\omega_i)$                  \\
        $\mathcal{A}_i$           & The $i^{th}$ amplitude of a spectrum                               \\
        $\Delta \mathcal{A}_i$    & The shift variable adding to normal spectrum\\
        $\Delta A$                & The expectation of shift variable \\
        $\alpha_i$                & The $i^{th}$ element in convolution kernel                         \\
        $\mathcal{F}_{i,j}$       & The Fourier result of $j^{th}$ base of $i^{th}$ feature            \\
        $\omega_{i,j}$            & The frequency of Fourier base corresponding to $\mathcal{F}_{i,j}$ \\
        $\boldsymbol{\Sigma}$     & The correlation matrix of joint distribution of amplitudes         \\
        $\nu_i$                   & The $i^{th}$ row and $i^{th}$ column of $\boldsymbol{\Sigma}$      \\
        $m$                       & The number of feature dimensions                                   \\
        $n$                       & The number of signals in a spectrum                                \\
        $k$                       & The number of signals in the context-aware selecting subset        \\
        $\boldsymbol{\mu}$        & The expectation of joint distribution of amplitudes                \\
        $\mu_i$                   & The $i^{th}$ element of $\boldsymbol{\mu}$                         \\ \hline
    \end{tabular}%
\end{table}

\section{Proposed Method}
\label{seg:method}
\begin{figure}[tbhp]
    \centering
    \includegraphics[width=\linewidth]{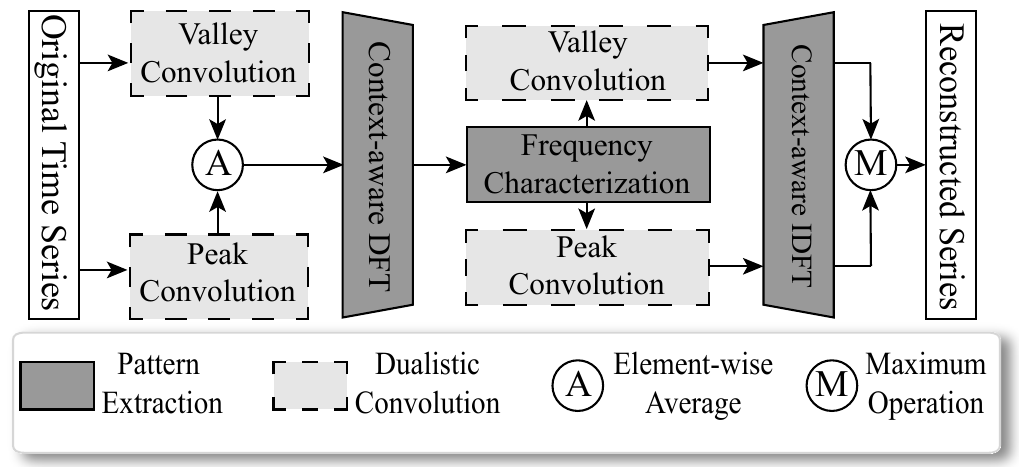}
    \caption{The model architecture of MACE}
    \label{fig:modelArch}
\end{figure}
\subsection{Overview}
The overview of MACE is depicted in Figure \ref{fig:modelArch}. There are roughly four stages in MACE. In the first stage, it amplifies anomalies and makes them easier to detect. In the second stage, it extracts the normal pattern subspace of each service, transforms the time series into the frequency domain, and obtains its frequency representation. In the third stage, it reconstructs the representation in the frequency domain. In the final stage, it transforms the reconstructed spectrum back into the time domain. More specifically, the details are provided in the following:
\begin{itemize}
    \item \textbf{Stage 1.} \emph{Amplify anomalies:} MACE initially employs \emph{dualistic convolution} in the time domain to amplify anomalies. The dualistic convolution is armed with a peak convolution and a valley convolution, which are used to emphasize the upward deviations and downward deviations respectively. We average the results of peak convolution and valley convolution in an element-wise manner.
    \item \textbf{Stage 2.} \emph{Time domain $\rightarrow$ Frequency representation:}  We present a new \emph{pattern extraction} mechanism to leverage the sparsity inherent in the frequency domain, thereby augmenting the model's capacity for generalization across various normal patterns. The \emph{pattern extraction} mechanism consists of a context-aware Discrete Fourier Transform (DFT), a frequency characterization module and a context-aware Inverse Discrete Fourier Transform (IDFT). The context-aware DFT identifies a normal pattern subspace in the frequency domain by selecting a subset of Fourier bases for each service and projects the services' data sample to the subspace in the frequency domain. After that, the frequency characterization module learns a frequency representation for the sample. \textcolor{black}{In this way, we learn a correlation between the data sample and a normal frequency subspace. In the following, we detect anomalies from this correlation instead of the data sample itself.}
    \item \textbf{Stage 3.} \emph{Reconstruction:} MACE replaces the vanilla convolution in the auto-encoder with the peak convolution and valley convolution separately to reconstruct the frequency representation, which enlarges the reconstruction error disparity between normality and anomalous.
    \item \textbf{Stage 4.} \emph{Reconstructed spectrum $\rightarrow$ Time domain:} MACE uses the context-aware IDFT in pattern extraction to transform the spectrums reconstructed by peak convolution and valley convolution back to the time domain. Finally, it selects the time series with the highest reconstruction error as the final reconstructed time series.
\end{itemize}

\textcolor{black}{During the training process, we use stochastic gradient descent to minimize the reconstruction error obtained from stage 4, which requires determining a hyperparameter -- learning rate.}

\textcolor{black}{
    There may be a concern about why we use different operations in stage 1 and stage 2 to the results of peak and valley convolution. It is worth noting that in stage 1, what we average is the manipulated data, while in stage 4, from what we select the maximum is the reconstruction error. In stage 1, we use average operation. In this way, the amplified downward and upward deviations are fused into the following networks. In stage 4, since the peak convolution and valley convolution enlarge the reconstruction error of upward deviations and downward deviations separately, for each time slot we compare the reconstruction error of them and pick the maximum one. In this way, we can detect both the upward deviation and the downward deviation. } 

\subsection{Dualistic Convolution}
\label{sec:dual}
The dualistic convolution consists of a peak convolution and a valley convolution, which targets at upward deviation and downward deviation respectively. It exhibits different effects in the time domain and frequency domain. 
In the time domain, it is reported that the short-term anomalies are easily neglected by encoder-decoder models \cite{DBLP:journals/pvldb/TuliCJ22}. Thus, we propose the dualistic convolution mechanism to extend the anomaly in the time domain, which makes the anomalies more conspicuous and easily detected. In the frequency domain, it hinders the reconstruction of anomalies, while keeping the reconstruction of normal samples easy to facilitate the model to identify anomalies. 

\begin{figure*}[t]
    \centering 
    \subfigure[An example for peak convolution.]{
    \includegraphics[width=0.3\linewidth]{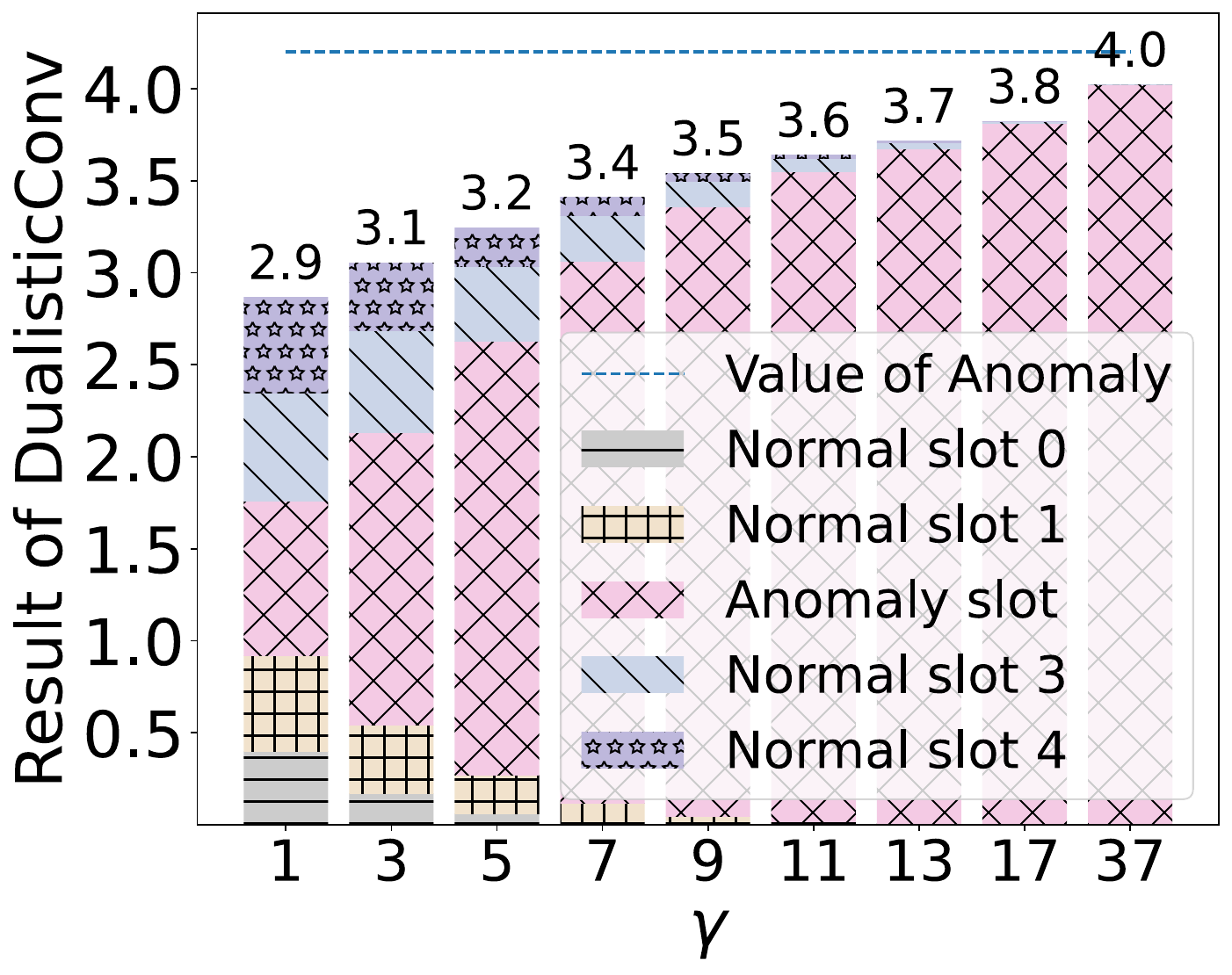}
    \label{fig:peak}
    }
    \hfill
    \subfigure[The dualistic and standard convolution in the time domain.]{
    \includegraphics[width=0.3\linewidth]{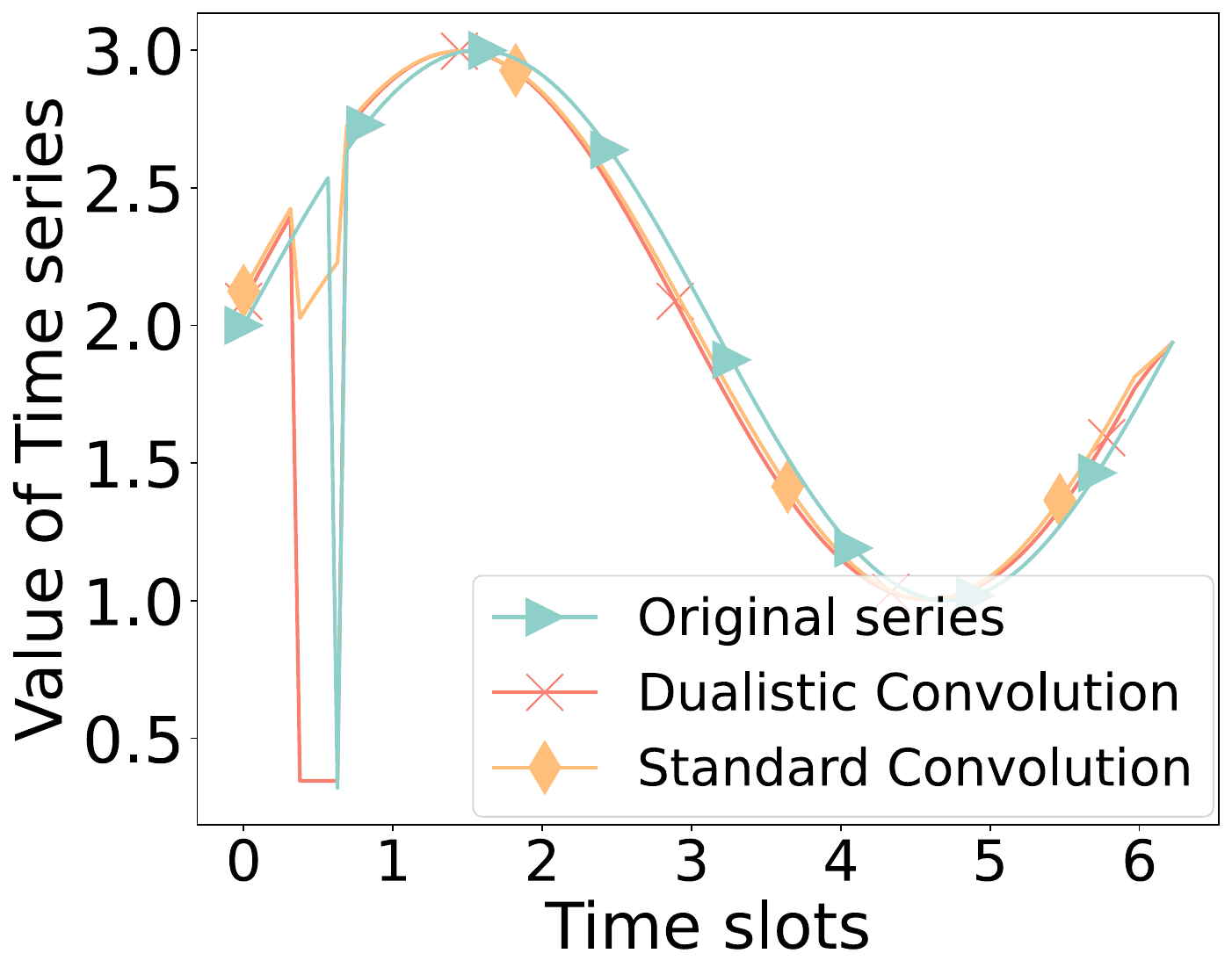}
    \label{fig:ampl2}
    }
    \hfill
    \subfigure[\textcolor{black}{The dualistic and standard convolution in the frequency domain.}]{
    \includegraphics[width=0.3\linewidth]{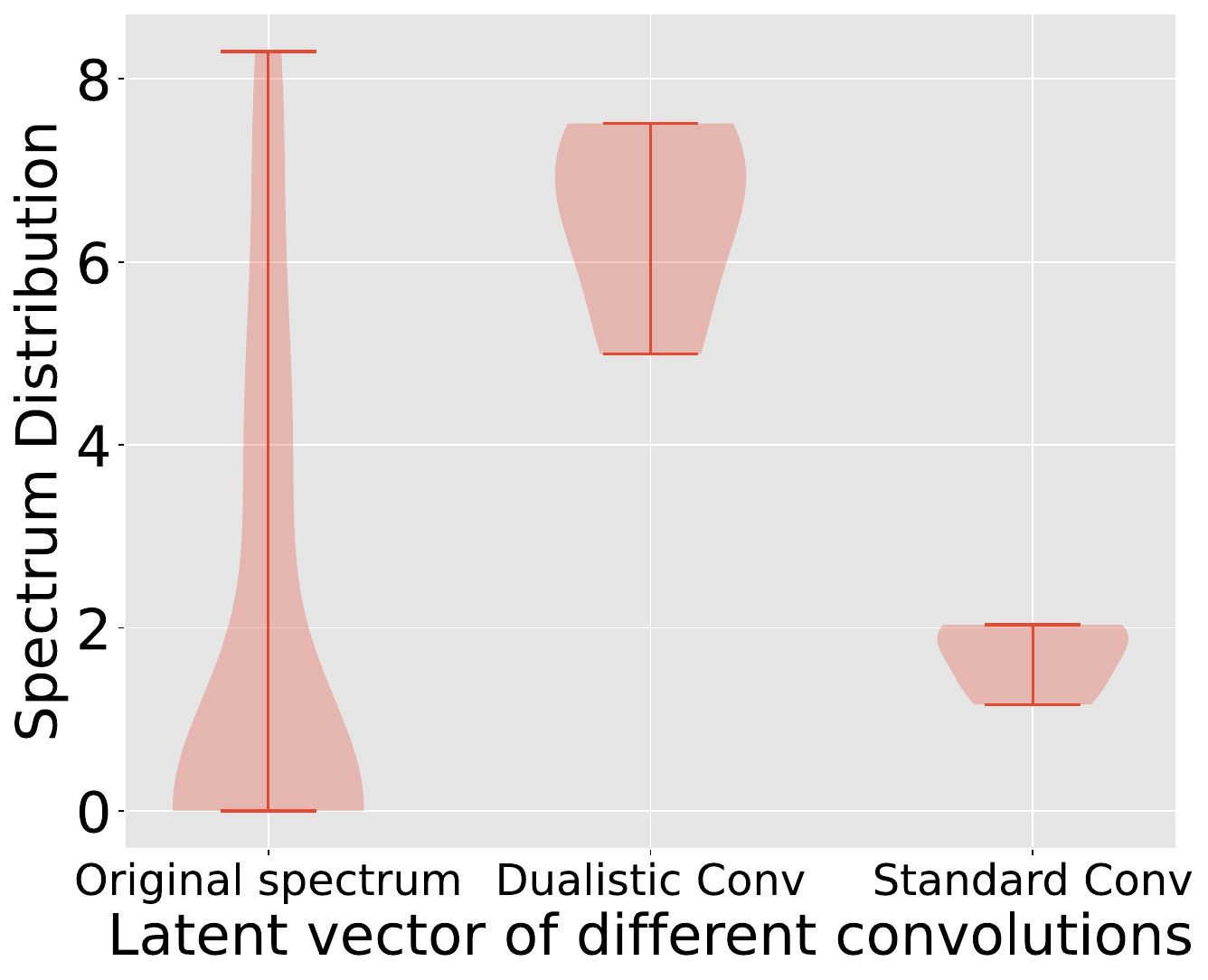}
    \label{fig:recon}
    }
    \hfill
    \caption{\label{fig:convVisual}(a) The figure shows the contributions of different time slots in peak convolution result in a convolution window when specifying different $\gamma$. As $\gamma$ grows, the contribution of deviations increases significantly. (b)-(c) The utility of dualistic convolution, compared with standard convolution.}
\end{figure*}

The dualistic convolution mechanism is depicted in Eq.\ref{eq:dual}, where $\gamma$ is a hyperparameter to make the convolution pay more attention to the deviations, $\sigma$ is a scaling hyperparameter and $\operatorname{Conv}_{1\times L}(x,s)$ denotes applying standard convolution to $x$ with stride $s$ and kernel length $L$. The peak convolution and valley convolution are obtained by setting different $\gamma$. 
The peak convolution is the dualistic convolution with $\gamma$ greater than $1$, while the valley convolution is the dualistic convolution with $\gamma$ less than $-1$. The larger the absolute value of $\gamma$ is, the more dominant the deviations are in the convolution result, as an example shown in Fig.\ref{fig:peak}. 
\begin{equation}
    \label{eq:dual}
    \begin{split}
    \operatorname{DualisticConv}(x) & =\sqrt[\gamma]{\operatorname{Conv}_{1\times L}(\frac{x^\gamma}{\sigma}, s)}\\
     \gamma  \in \{2g+1| g \in & Z\  \land \ g \neq 0  \},\  \sigma  > 0
    \end{split}
\end{equation}

The dualistic convolution yields distinct outcomes in both the time and frequency domains. To illustrate this visually, the convolution results are depicted in Fig.\ref{fig:convVisual}. In the time domain, standard convolution mitigates deviations, while dualistic convolution amplifies them. In the frequency domain, standard convolution compresses the spectrum to a low-dimensional space around the main body of the spectrum distribution, whereas dualistic convolution compresses it to the space around the "long tail" of the spectrum distribution. As confirmed later, anomalous data typically exhibits a long-tail distribution in the frequency spectrum, while normal data rarely shows such a tail. Considering that reconstruction is easier from latent vectors close to the main body of the distribution than from those close to the tail, dualistic convolution hinders the reconstruction of anomalies while keeping the process for normal data straightforward.

\textbf{In the Time Domain,} the stride $s$ of dualistic convolution is set to 1. In this way, the dualistic convolution functions as a weighted summation operator for each kernel-length sliding window, which emphasizes the deviations in its results. Thus, once an anomaly is included in a convolution window, the convolution result of this window will be dominated by the anomaly. Consequently, short-term anomaly will be extended by the kernel length, as shown in Fig.\ref{fig:ampl2}, which makes short-term anomaly easier to detect.    

\begin{figure}[t]
    \centering 
    \subfigure[Dualistic convolution in the frequency domain.]{
    \includegraphics[width=0.6\linewidth]{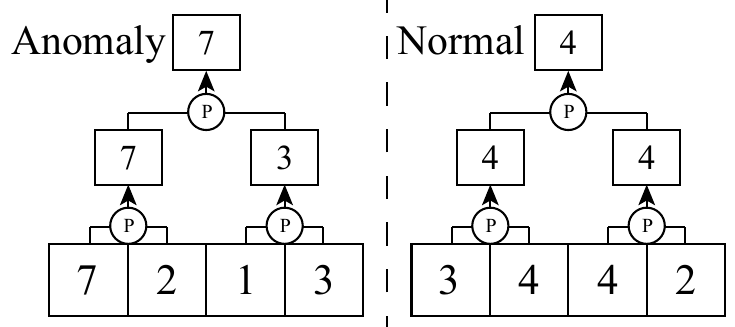}
    \label{fig:freqConv}
    }
    \hfill
    \subfigure[The three channels of frequency representation.]{
    \includegraphics[width=0.3\linewidth]{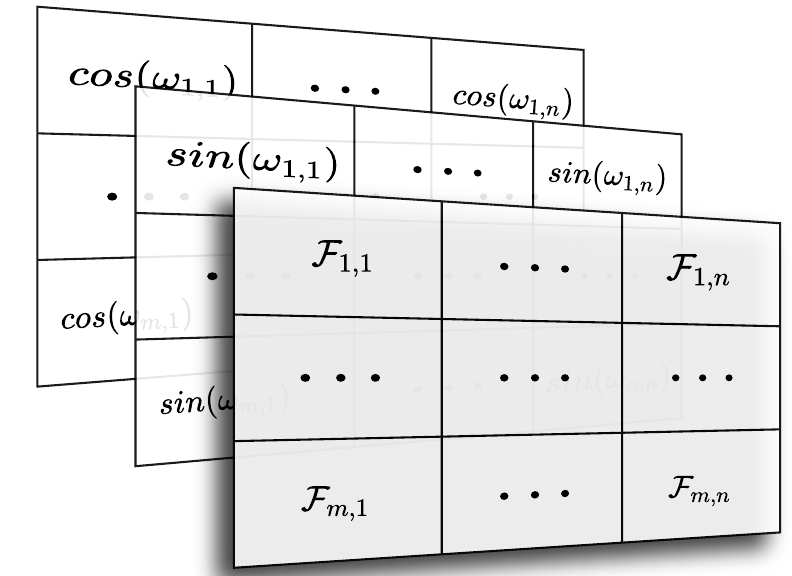}
    \label{fig:freqEmb}
    }
    \caption{(a) The dualistic convolution applying to the frequency domain actually picks the prominent deviation in each compression step. (b) The figure shows the three channels of frequency representation in the frequency characterization module. The first channel is the result of Fourier transformation, the second is corresponding $sin$ Fourier bases and the third is corresponding $cos$ Fourier bases.}
\end{figure}

\textbf{In the Frequency Domain,} it has been reported that most anomalies manifest themselves as strong signals with high-energy components \cite{ma2021jump}, which renders their spectrum higher variability. This can be proven by statistical data in Table \ref{Tab:var}, where the amplitude variances of anomalies are greater than those of normal patterns. Consequently, we set the stride of dualistic convolution in the frequency domain to the size of the convolution kernel. In this way, the dualistic convolution actually picks the min (valley convolution) or the max (peak convolution) amplitude in each kernel-length segment to comprise the latent vector, as shown in Fig.\ref{fig:freqConv}. Intuitively, the dualistic convolution in the frequency domain hinders the reconstruction of anomalies and keeps the reconstruction of normalities easy, because when the components in a spectrum are highly variable, the dualistic convolution tends to pick the high-energy components to comprise its latent vector, which can obviously deviate from other components and are difficultly reconstructed from. In contrast, when the components in a spectrum are closer to each other, the latent vector obtained by dualistic convolution does not deviate significantly from the original spectrum and can represent it better. Furthermore, we conduct a theoretical comparison of the challenges involved in reconstructing a spectrum from its latent vector for both normal and anomalous cases. The level of reconstruction difficulty is directly associated with the gap between the latent vector and the original spectrum. 
In Theorem 1, we examine the upper bound of this gap for both normal and anomalous scenarios. Our analysis reveals that the constraints on the gap in normal cases are more stringent when compared to those in anomalous cases.

\emph{Definition 1.} 
The gap between the convolution result and the original spectrum of each convolution window is defined as \textcolor{black}{$\sum_{j=1}^n \mathbb{E}(\left |\operatorname{DualisticConv(\mathcal{A})} - \mathcal{A}_j \right |)$}, where $\mathcal{A}$ is the amplitudes of spectrum in the convolution window, $\mathcal{A}_j$ is the $j^{th}$ element in $\mathcal{A}$ and $n$ is the kernel length.

\emph{Theorem 1.} When the amplitudes follow a Gaussian joint distribution $\mathcal{N}(\boldsymbol{\mu}, \boldsymbol{\Sigma})$, the distance between the latent vector and the original spectrum is upper bounded by $2^{\frac{\gamma-1}{\gamma}}n\sqrt[\gamma]{\sum_{i=1}^n (\gamma-1)!!\nu_i^\gamma\vert\alpha_i\vert+\vert\alpha_i \mu_i^{\gamma}\vert}-\sum_{j=1}^n\mu_j $, where $\alpha_i$ is the $i^{th}$ element in the kernel divided by $\sigma$, $\mu_i$ is the $i^{th}$ element of $\boldsymbol{\mu}$, $\nu_i$ is the $i^{th}$ row and $i^{th}$ column element of $\boldsymbol{\Sigma}$ and $n!!$ denotes $[n (n-2) (n-4) \cdots 1]$.

\emph{Proof skeleton.} \textcolor{black}{We first transform the gap formation to $\left | \sum_{j=1}^n \mathbb{E}(\operatorname{DualisticConv(\mathcal{A})} - \mathcal{A}_j) \right |$. That is because $\operatorname{DualisticConv}(\mathcal{A})-\mathcal{A}_j \geq 0, \forall j$ as long as we choose a big enough $\gamma$. Taking the peak convolution as an example,  $\lim_{\gamma \rightarrow \inf}\operatorname{DualisticConv}(\mathcal{A})=\max(\mathcal{A}_1,\mathcal{A}_2,\dots, \mathcal{A}_n)$. Thus, when we take a big $\gamma$, we can approximately confirm that $\mathcal{A}-\mathcal{A}_j \geq 0, \forall j$.} Subsequently, we further transform the gap as shown in Eq.\ref{pf1}-Eq.\ref{pf2}. Since the function $f(x)=\sqrt[\gamma]{x}$ is a concave function when $x \geq 0$, it can be further scaled by Jensen inequality \cite{kim2016further} as shown in Eq.\ref{pf3}. Let $\mathcal{A}_i=\bar{\mathcal{A}}_i+\mu_i$, where $\bar{\mathcal{A}}_i \sim \mathcal{N}(\boldsymbol{0},\boldsymbol{\Sigma})$, and we obtain Eq.\ref{pf4}. The equation is further scaled by power mean inequality \cite{wang2014power} and we obtain Eq.\ref{pf5}. Facilitated by the property of Gamma Function \cite{artin2015gamma}, it can be computed that $\mathbb{E}(\vert\bar{\mathcal{A}}_i^{\gamma}\vert)=(\gamma-1)!!\nu_i^\gamma$. Thus, we get the conclusion, as shown in Eq.\ref{pf6}.
\textcolor{black}{
\begin{gather}
    \hspace{-3.5cm} \left | \sum_{j=1}^n  \mathbb{E}(\operatorname{DualisticConv(\mathcal{A})}  - \mathcal{A}_j) \right | \\
   \hspace{-2.99cm}= \left | \mathbb{E}(n\sqrt[\gamma]{\sum_{i=1}^n \alpha_i \mathcal{A}_i^{\gamma}}) -\sum_{j=1}^n\mu_j  \right | \label{pf1} \\
   \hspace{-2.58cm} \leq \left | \mathbb{E}(n\sqrt[\gamma]{\sum_{i=1}^n \vert\alpha_i\vert \vert\mathcal{A}_i^{\gamma}\vert}) -\sum_{j=1}^n\mu_j \label{pf2} \right | \\
   \hspace{-2.58cm} \leq \left | n\sqrt[\gamma]{\sum_{i=1}^n \vert\alpha_i\vert \mathbb{E}(\vert\mathcal{A}_i^{\gamma}\vert)} -\sum_{j=1}^n\mu_j \label{pf3} \right | \\
   \hspace{-1.4cm} = \left | n\sqrt[\gamma]{\sum_{i=1}^n \vert\alpha_i\vert \mathbb{E}(\vert(\bar{\mathcal{A}}_i+\mu_i)^{\gamma}\vert)} -\sum_{j=1}^n\mu_j \label{pf4} \right | \\
   \hspace{-0.45cm} \leq \left | 2^{\frac{\gamma-1}{\gamma}}n\sqrt[\gamma]{\sum_{i=1}^n \vert\alpha_i\vert \mathbb{E}(\vert\bar{\mathcal{A}}_i^{\gamma}\vert)+\vert\alpha_i\mu_i^{\gamma}\vert} -\sum_{j=1}^n\mu_j \right | \label{pf5} \\
   \hspace{0.4cm} =\left |2^{\frac{\gamma-1}{\gamma}}n\sqrt[\gamma]{\sum_{i=1}^n \vert\alpha_i\vert (\gamma-1)!!\nu_i^\gamma+\vert\alpha_i\mu_i^{\gamma}\vert} -\sum_{j=1}^n\mu_j \right | \label{pf6}
\end{gather}
}
It is worth noting that the upper bound is primarily determined by $\nu_i, i\in [1,n]$, whereas the influence of $\boldsymbol{\mu}$ is negligible. This is because, regardless of the value of $\boldsymbol{\mu}$, the expression $2^{\frac{\gamma-1}{\gamma}}n\sqrt[\gamma]{\sum_{i=1}^n \vert\alpha_i\vert (\gamma-1)!!\nu_i^\gamma+\vert\alpha_i\mu_i^{\gamma}\vert} -\sum_{j=1}^n\mu_j$ is always greater than $2^{\frac{\gamma-1}{\gamma}}n\sqrt[\gamma]{\sum_{i=1}^n \vert\alpha_i\vert (\gamma-1)!!\nu_i^\gamma}$,  which is solely related to $\nu_i, i\in[1,n]$. This can be proven using the power mean inequality.
As a result, the upper bound for the gap is primarily influenced by the standard deviation of amplitudes and positively correlated with it. Consequently, the gap between the latent vector and the original amplitudes for normal distributions is more rigorously constrained, implying that they are easier to reconstruct.
\begin{table}[]
    \centering
    \renewcommand\arraystretch{1}
    \caption{\label{Tab:var}The average spectrum variances of anomalous and normal patterns in different datasets.}
    \begin{tabular}{l|l|l|l}
    \hline
            & SMD \tablefootnote{The Server Machine Dataset \cite{su2019robust}}  & J-D1 \tablefootnote{A service dataset from one of global top 10 internet company \cite{ma2021jump}}  & J-D2 \tablefootnote{A service dataset from one of global top 10 internet company \cite{ma2021jump}} \\ \hline
    Anomaly & 4.55 & 12.38 & 15.64 \\ \hline
    Normality  & 3.36 & 11.74 & 14.13 \\ \hline
    \end{tabular}%
\end{table}

\subsection{Pattern Extraction}
\label{sec:PaExtr}
The most challenging problem in tackling diverse normal patterns with a unified model is that an anomaly for one normal pattern can be normal for another. To overcome this issue, we propose a pattern extraction mechanism to detect anomalies by the correlation between the data sample and its service normal pattern, instead of the data sample itself, which allows us to handle various normal patterns and reduce computational overhead by capitalizing on the sparsity inherent in the frequency domain. Our pattern extraction mechanism comprises three key components: a context-aware discrete Fourier transformation (DFT) module, a frequency characterization module, and a context-aware inverse discrete Fourier transformation (IDFT) module.
In the preprocessing stage, we analyze each service in the frequency domain and identify a normal pattern subspace containing most of the normalities as the service normal pattern subspace by
establishing a compact set of dominant Fourier bases for each normal pattern. During both the training and testing phases, we employ the context-aware DFT module to project time series data to its service normal pattern subspace by approximating it with a linear combination of the bases within the relevant normal pattern subspace. This procedure effectively compresses the spectral volume and minimizes computational demands.
Subsequently, the frequency characterization module is used to create a tailored frequency representation for the time series data from its projection. After undergoing reconstruction through a dualistic convolution-based auto-encoder, the spectrum is transformed back into time series data using the context-aware IDFT. Furthermore, we theoretically prove that utilizing only the dominant bases for each normal pattern yields superior performance in distinguishing anomalies from normal patterns compared to using the complete spectrum. This is further supported by experimental evidence in Section \ref{sec:abl}.

\textbf{Context-aware DFT and IDFT.} It is assumed that each service or server exhibits its unique normal pattern. Consequently, during the preprocessing stage, we process the training dataset in the frequency domain and count the occurrences of each Fourier base as the first $k$ strongest signals across all sliding windows. Subsequently, we select the top $k$ bases with the highest incidence to serve as Fourier bases for their service normal patterns subspace.
In both the training and testing phases, the context-aware DFT transforms the time series data exclusively using the bases from the corresponding normal pattern subspace through the DFT process. Likewise, the context-aware IDFT processes the spectrum using only the corresponding bases through the IDFT. Furthermore, we conduct a theoretical comparison of the reconstruction error between anomalies and normal patterns, demonstrating that the context-aware DFT can significantly widen the gap in reconstruction errors between anomalies and normal patterns.

\emph{Definition 2} (Spectrum). Given a DFT spectrum of a normal pattern $\mathcal{A}_N(\omega_0)> \dots> \mathcal{A}_N(\omega_n)$, where $\mathcal{A}_N(\omega_i)$ is the amplitude of the signal and $\omega_i$ is its corresponding frequency, we compute normalized value of them as follows: $q_N(\omega_i)=\frac{\mathcal{A}_N(\omega_i)}{\sum_{i=1}^n \mathcal{A}_N(\omega_i)}$. The spectrum of anomalies is denoted by $\mathcal{A}_A(\omega_0), \dots, \mathcal{A}_A(\omega_n)$, where $\omega_i$ is exactly the $\omega_i$ in spectrum of normal pattern. Similarly, the spectrum of anomalies is normalized and denoted by $q_A(\omega_i)$. The normalized spectrum for normalities and anomalies obtained by context-aware DFT are denoted by $\bar{q}_N(\omega_i)$ and $\bar{q}_A(\omega_i)$ respectively.

\emph{Definition 3} (Reconstruction error). The reconstruction error of context-aware DFT is defined as the KL divergence between the spectrum obtained by context-aware DFT and the original spectrum, i.e. $\operatorname{KL}(\bar{q}|q)$.

\emph{Assumption 1.} The anomalies manifest themselves by adding a shift variable to the spectrum of normalities, whose expectation is greater than 0, i.e. $\mathcal{A}_A(\omega_i)=\mathcal{A}_N(\omega_i)+\Delta\mathcal{A}_i$, where $\forall i, \Delta\mathcal{A}_i$ are independent identically distributed and follow an unknown distribution with expectation $\Delta A$, $\Delta A>0$. 
It is reasonable to assume the expectation of shift variable is bigger than 0 since it is reported that the anomalies have stronger signals than normalities \cite{ma2021jump}, which implies higher amplitude expectations of anomalies. Moreover, we statistically collect the expectation of anomalies and normalities across three real-world datasets and verify this point, as shown in Table \ref{Tab:mean}.

\begin{table}[]
    \centering
    \vspace{-3mm}
    \renewcommand\arraystretch{1}
    \caption{\label{Tab:mean}The expectation of amplitudes for anomalies and normalities on different datasets.}
    \begin{tabular}{l|l|l|l}
    \hline
            & SMD  & J-D1 & J-D2 \\ \hline
    Anomaly & 0.36 & 0.74 & 0.81 \\ \hline
    Normality  & 0.23 & 0.72 & 0.77 \\ \hline
    \end{tabular}%
    \vspace{-3mm}
\end{table}

\emph{Theorem 2.} The gap of reconstruction error between the anomaly and normality is $\log \frac{\sum_{i=1}^k q_N(\omega_i)}{\sum_{i=1}^k q_A(\omega_i)}$.

\emph{Proof.} 
Using normality as an example, we derive the expression for its reconstruction error. The expression for an anomaly can be derived in a similar manner. We start by representing $\bar{q}_N$ as shown in Eq. \ref{eq:pf2-1}. Subsequently, the reconstruction error for it is obtained in Eq. \ref{eq:pf2-2}. Therefore, the difference in the reconstruction error between anomalies and normal patterns is given by $\operatorname{KL}(\bar{q}_A|q_A)-\operatorname{KL}(\bar{q}_N|q_N)=\log \frac{\sum_{i=1}^k q_N(\omega_i)}{\sum_{i=1}^k q_A(\omega_i)}$.

\begin{equation}
    \label{eq:pf2-1}
    \bar{q}_N(\omega_i)=\left\{\begin{matrix}
        \frac{q_N(\omega_i)}{\sum_{i=1}^kq_N(\omega_i)} &, i\leq k\\
        0 &, i > k
    \end{matrix}\right.
\end{equation}

\begin{gather}
    \label{eq:pf2-2}
    \begin{split}
    \operatorname{KL}(\bar{q}_N|q_N) & =\sum_{j=1}^k\frac{q_N(\omega_j)}{\sum_{i=1}^kq_N(\omega_i)}\log\frac{1}{\sum_{t=1}^kq_N(\omega_t)}  \\
    & =-\frac{\sum_{j=1}^kq_N(\omega_j)}{\sum_{t=1}^kq_N(\omega_t)}\log\sum_{i=1}^kq_N(\omega_i)  \\
    & =-\log\sum_{i=1}^kq_N(\omega_i)
    \end{split}
\end{gather}
Intuitively, the gap is greater than 0 because the numerator $\sum_{i=1}^k q_N(\omega_i)$ represents the first $k$ strongest signals, while the denominator $\sum_{i=1}^k q_A(\omega_i)$ is not guaranteed to have a similar characteristic. We have conducted a further analysis of the condition for achieving a smaller reconstruction error for normal patterns in Corollary 1.

\emph{Corollary 1.} When $\sum_{i=1}^kq_N(\omega_i)>\frac{k}{n}$, the reconstruction error of normality is smaller than that of anomaly.

\emph{Proof.} According to assumption 1, the $q_A(\omega_i)$ can be transformed into $\frac{q_N(\omega_i)S+{\Delta \mathcal{A}_i}}{S+{\sum_{i=1}^n\Delta \mathcal{A}_i}}$, where $S=\sum_{i=1}^n\mathcal{A}_N(\omega_i)$. Thus, the gap of reconstruction error between anomaly and normality can be transformed to $\log\frac{S+n\Delta A}{S+k \Delta A (\sum_{j=1}^kq_N(\omega_j))^{-1}} $ according to the Law of Large Numbers \cite{hsu1947complete}. When $\sum_{i=1}^kq_N(\omega_i)>\frac{k}{n}$, we can obtain $S+n\Delta A>S+k \Delta A (\sum_{j=1}^kq_N(\omega_j))^{-1}$. As a result, the gap of reconstruction error is bigger than 0.

It is worth noticing that when using trivial DFT and completed spectrum, $k$ is set to $n$ and the reconstruction error gap of anomaly and normality becomes zero, considering that $q_A$ and $q_N$ are normalized values. In contrast, by examining the condition in Corollary 1, it becomes evident that there must be a value of $k$ less than $n$ that yields a reconstruction error gap greater than 0. Thus, compared with using the full spectrum, only using a subset of them widens the reconstruction error gap between the normalities and anomalies. This confirms that anomalies are easier to distinguish when employing the context-aware DFT with a reduced set of bases compared to the standard DFT.

\textbf{Frequency characterization.} The frequency characterization module concatenates the result of context-aware DFT and explicit marked trigonometric bases, as shown in Fig.\ref{fig:freqEmb}, where $\mathcal{F}_{i,j}$ denotes the Fourier result of $j^{th}$ base for $i^{th}$ feature dimension, $\omega_{i,j}$ denotes the frequency of $j^{th}$ base for $i^{th}$ feature dimension. Afterward, we use a three-channel convolution to manipulate the concatenated tensors and obtain the frequency representation.

\section{Experiment}
\label{seg:exp}
We conduct extensive experiments on four real-world anomaly detection datasets and obtain the following conclusions:
\begin{itemize}
    \item When detecting anomalies for multiple services with kinds of normal patterns by a unified model, MACE achieves better performance compared with the state-of-the-art methods.
    \item MACE achieves comparable performance compared with the state-of-the-art methods when the state-of-the-art methods are trained separately for each service and MACE uses a unified model for all the services.
    \item MACE shows good transferability on unseen datasets.
    \item MACE consumes obviously less time and memory overhead than the state-of-the-art methods. 
    \item Every module in MACE contributes to its performance.
\end{itemize}
\subsection{Experiment Setup}
The datasets used in this paper contain several subsets, which represent data for different services, servers and detecting sensors. It is assumed that different subsets have different normal patterns. We divide every ten subsets in a dataset as a group. For each group, we train a unified model to detect the anomaly in it.  

\begin{figure*}[t]
    \centering 
    \vspace{-3mm}
    \subfigure[\textcolor{black}{The distribution of pairwise distance between subsets}]{
    \includegraphics[width=0.3\linewidth]{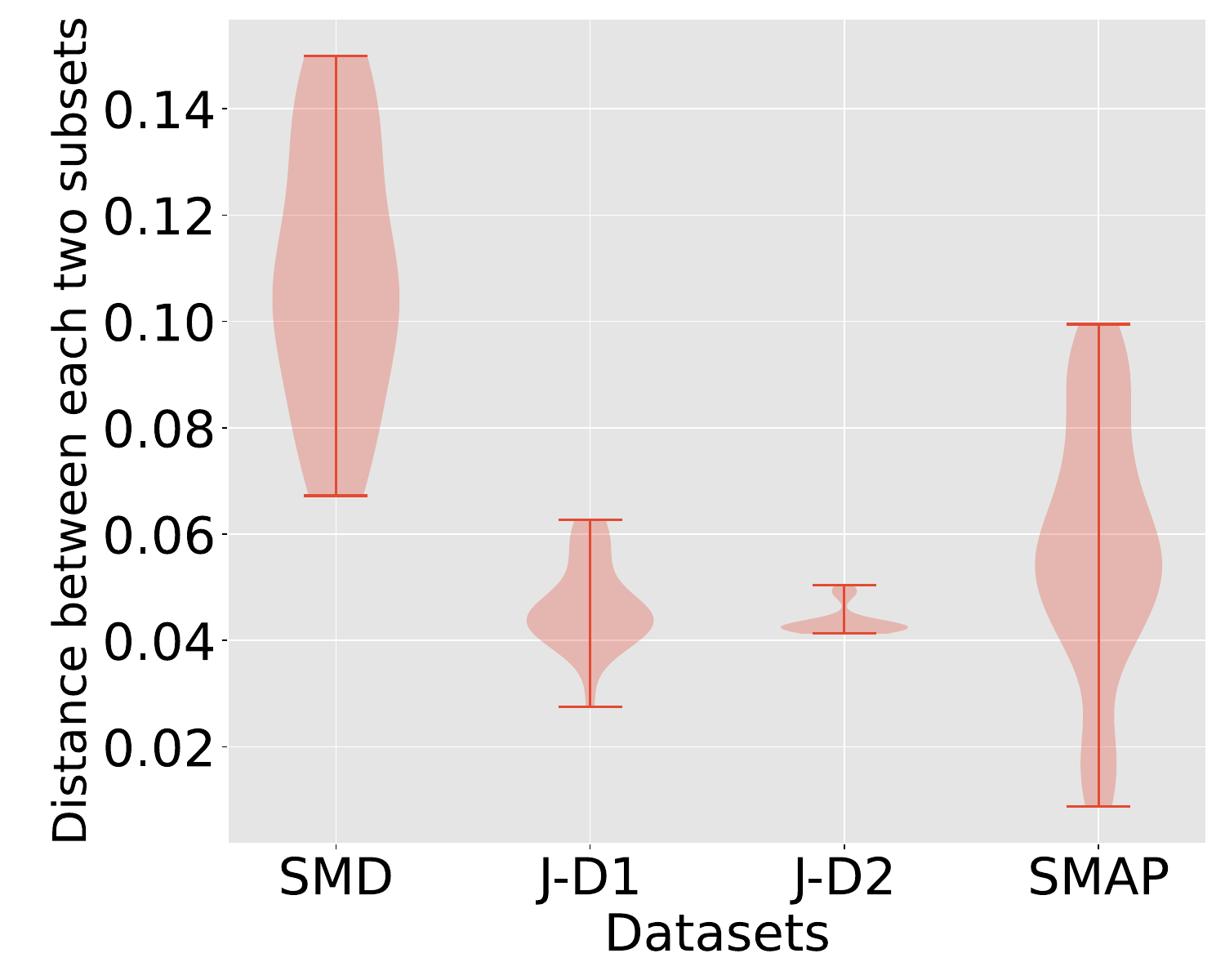}
    \label{fig:distance}
    }
    \hfill
    \subfigure[The ratios of point anomaly, context anomaly and normal patterns in each dataset]{
    \includegraphics[width=0.3\linewidth]{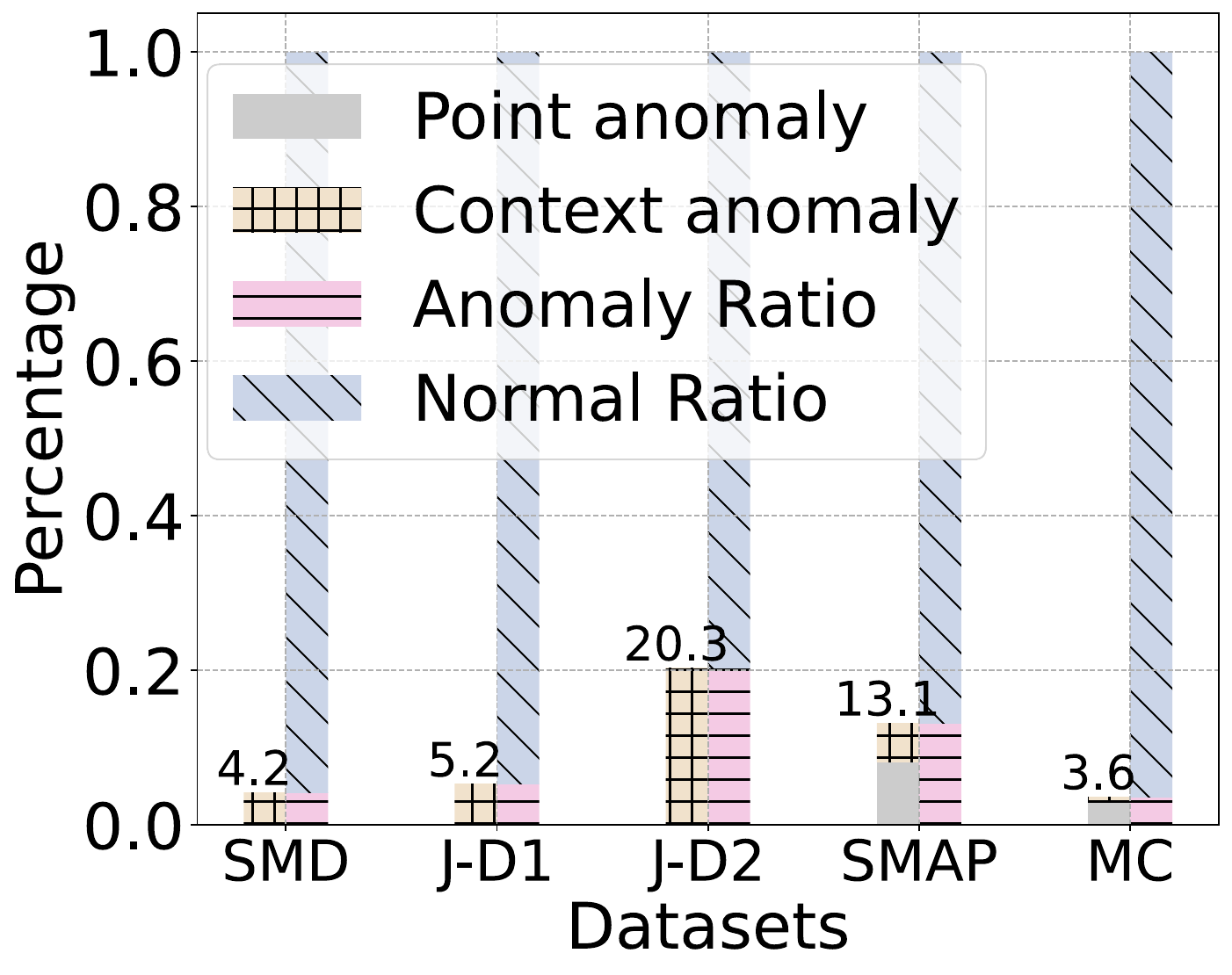}
    \label{fig:consist}
    }
    \hfill
    \subfigure[\textcolor{black}{The F1 score of unified model across various services}]{
    \includegraphics[width=0.28\linewidth]{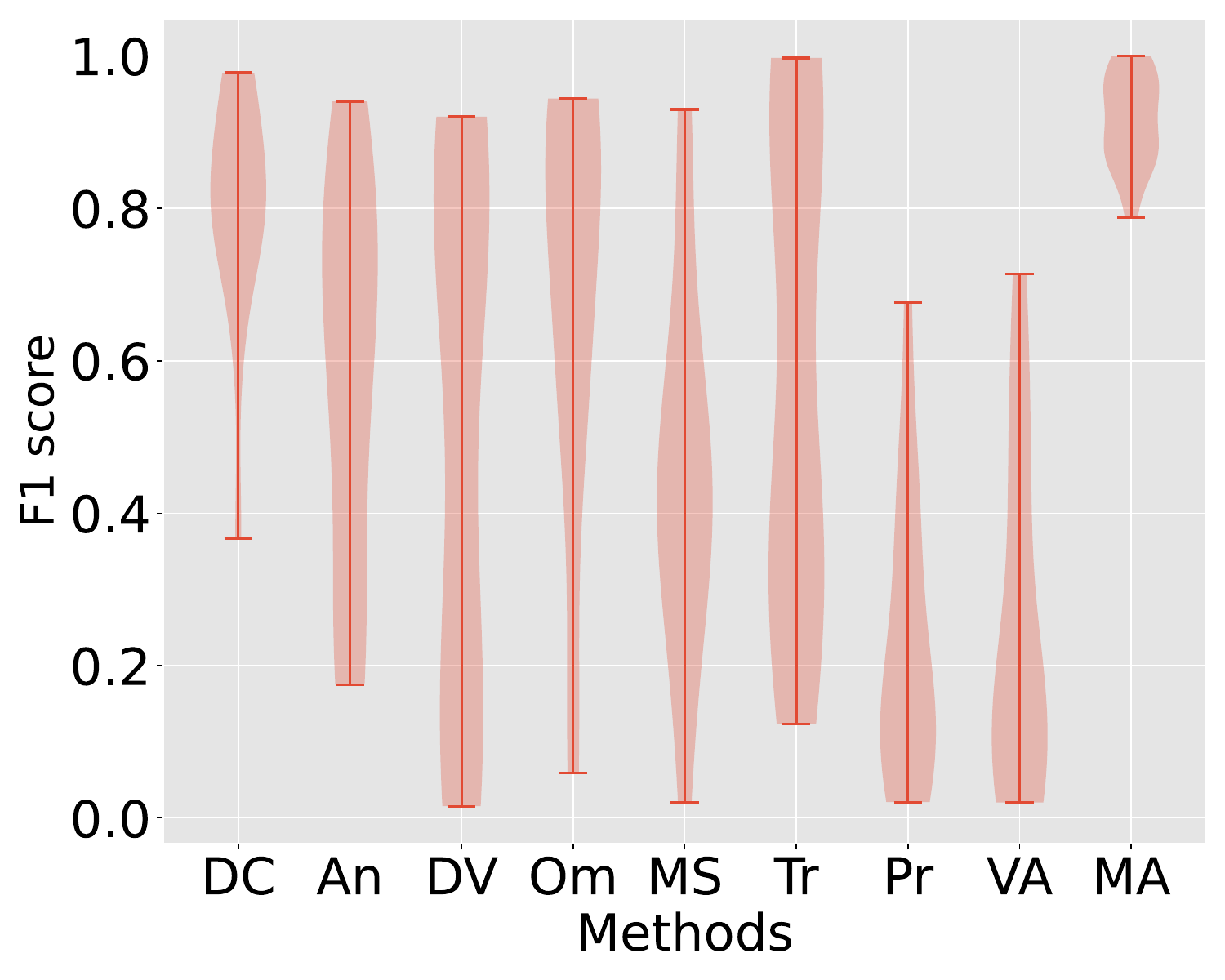}
    \label{fig:multi}
    }
    \hfill
    \caption{\label{Fig:datasetDiscript} (a) We first use kernel density estimation to estimate the distribution of each subset. Subsequently, we compute KL divergence between each pair of subsets in a training group. The figure shows the distribution of the KL divergences of different datasets. (b) The figure shows the point anomaly, context anomaly and normal pattern ratios in each dataset. (c) All the methods train a unified model for every ten services. The figure shows their F1 score across different services.}
    \vspace{-3mm}
\end{figure*}

\textbf{Datasets.} We utilized a selection of datasets, including the widely-recognized Server Machine Dataset (SMD), two cloud service monitoring logs from a top global Internet company (J-D1 and J-D2), the well-established anomaly detection benchmark, Soil Moisture Active Passive (SMAP) and a dataset MC that we collect from one of top ten global cloud providers. As shown in Fig.\ref{Fig:datasetDiscript}, the normal patterns of SMD are the most diverse, while the normal patterns of J-D2 are the most similar. Moreover, SMAP has most point anomalies, while the anomalies in other datasets are lasting. 
\begin{itemize}
    \item Server Machine Dataset (SMD) \cite{su2019robust}: SMD spans a 5-week period and originates from a major Internet company, incorporating data from 28 distinct machines. Each machine's log data, a subset of SMD, is equally divided into training and testing sets. The anomaly ratio in SMD is 4.16\%.
    \item Datasets provided by Jumpstarter (J-D1 and J-D2) \cite{ma2021jump}: J-D1 and J-D2 are two datasets gathered from a top global Internet company. Each dataset includes logs of 19 metrics from 30 services, with each service's log data forming a subset in J-D1 and J-D2. The anomaly ratios for J-D1 and J-D2 are 5.25\% and 20.26\%, respectively.
    \item Soil Moisture Active Passive (SMAP) \cite{hundman2018detecting}: SMAP comprises real spacecraft telemetry data and anomalies from the Soil Moisture Active Passive satellite, featuring an anomaly ratio of 13.13\%.
    \item {MC: MC consists of the monitoring data from 25 services for 15 days, whose anomaly ratio is 3.6\% and contains substantial point anomalies.}
\end{itemize}

\textbf{Hyperparameter.} 
The important hyperparameters of MACE are shown in Table \ref{Tab:hyper}, where $m$ denotes the number of basis in a subset, $\gamma_f$ denotes $\gamma$ for the dualistic convolution in the frequency domain, $\gamma_t$ denotes the one in the time domain, $\sigma_f$ denotes the scaling factor of dualistic convolution in the frequency domain and $\sigma_t$ denotes the one in the time domain.
\begin{table}[htbp]
    \centering
    \vspace{-3mm}
    \renewcommand\arraystretch{1}
    \caption{\label{Tab:hyper}Hyperparameters of MACE.}
    \begin{tabular}{cc|cc}
        \hline
        \textbf{Hyperparameter} & \textbf{Value} & \textbf{Hyperparameter} & \textbf{Value} \\ \hline
        $m$                     & 20             & $\gamma_f$ in SMD       & 7              \\
        $\gamma_t$ in SMD       & 11             & $\gamma_f$ in J-D1      & 11             \\
        $\gamma_t$ in J-D1      & 11             & $\gamma_f$ in J-D2      & 13             \\
        $\gamma_t$ in J-D2      & 13             & $\gamma_f$ in SMAP      & 13             \\
        $\gamma_t$ in SMAP      & 13             & $\sigma_f$ in SMD       & 5              \\
        $\sigma_t$ in SMD       & 5              & $\sigma_f$ in J-D1      & 5              \\
        $\sigma_t$ in J-D1      & 5              & $\sigma_f$ in J-D2      & 5              \\
        $\sigma_t$ in J-D2      & 5              & $\sigma_f$ in SMAP      & 5              \\
        $\sigma_t$ in SMAP      & 7              & kernel length           & 5              \\
        window size             & 40             & learning rate           & 0.001          \\ \hline
    \end{tabular}%
    \vspace{-3mm}
\end{table}

\textbf{Baselines.}
We conducted a comprehensive comparison of MACE with several state-of-the-art methods, including DCdetector \cite{DBLP:conf/kdd/YangZZW023}, AnomalyTransformer \cite{DBLP:conf/iclr/XuWWL22}, DVGCRN \cite{DBLP:conf/icml/ChenTCDDZ22}, JumpStarter \cite{ma2021jump}, OmniAnomaly \cite{su2019robust}, and MSCRED \cite{zhang2019deep}. To assess its diverse pattern generalization capabilities, we introduced two additional methods: TranAD, a meta-learning-based approach \cite{DBLP:journals/pvldb/TuliCJ22}, and ProS, a transfer-learning-based method \cite{kumagai2019transfer}. Furthermore, to evaluate its computational efficiency, we compared MACE with the classical anomaly detection method VAE \cite{DBLP:journals/corr/KingmaW13}. Due to space constraints, some figures represent methods using the first two letters of their names as a shorthand.
\begin{itemize}
    \item DCdetector (DC) \cite{DBLP:conf/kdd/YangZZW023}: DCdetector is a latest cutting-edge anomaly detection method. It employs a unique dual attention asymmetric design to establish a permuted environment and leverages pure contrastive loss to guide the learning process. This enables the model to learn a permutation-invariant representation with superior discrimination abilities.
    \item AnomalyTransformer (An) \cite{DBLP:conf/iclr/XuWWL22}:  Anomaly Transformer stands out as one of the most cutting-edge methods, harnessing the formidable capabilities of transformers to model point-wise representation and pair-wise associations through an innovative anomaly-attention mechanism.
    \item DVGCRN (DV) \cite{DBLP:conf/icml/ChenTCDDZ22}: DVGCRN stands out as another cutting-edge anomaly detection method, effectively modeling fine-grained spatial and temporal correlations in multivariate time series. It achieves a precise posterior approximation of latent variables, contributing to a robust representation of multivariate time series data.
    \item JumpStarter (Ju) \cite{ma2021jump}: JumpStarter stands out as a cutting-edge anomaly detection method, equipped with a shape-based clustering and an outlier-resistant sampling algorithm. This combination ensures a rapid initialization and high F1 score performance. 
    \item OmniAnomaly (Om) \cite{su2019robust}: OmniAnomaly is a widely acknowledged anomaly detection method, employing a stochastic variable connection and planar normalizing flow to robustly capture the representation of normal multivariate time series data.
    \item MSCRED (MS) \cite{zhang2019deep}: MSCRED is a highly acclaimed method known for its ability to detect anomalies across various scales and pinpoint root causes through the utilization of multi-scale signature matrices.
    \item TranAD (Tr) \cite{DBLP:journals/pvldb/TuliCJ22}: TranAD, a meta-learning-based anomaly detection method, represents one of the latest cutting-edge approaches. It excels in learning a robust initialization for anomaly detection models, demonstrating excellent generalization across diverse normal patterns.
    \item ProS (Pr) \cite{kumagai2019transfer}: ProS introduces a zero-shot methodology, capable of inferring in the target domains without the need for re-training. This is achieved through the introduction of latent domain vectors, serving as latent representations of the domains.
    \item VAE (VA) \cite{DBLP:journals/corr/KingmaW13}: VAE, a widely recognized and classical anomaly detection method, serves as a foundational framework for numerous state-of-the-art approaches. It introduces low computational and memory overhead, contributing to its popularity.
\end{itemize} 

\begin{table*}[htbp]
    \centering
    \vspace{-3mm}
    \renewcommand\arraystretch{1}
    \caption{\label{Tab:pq1}The performance of MACE and baselines when detecting anomalies for multiple patterns by a unified model. }
    \begin{tabular}{l|ccc|ccc|ccc|ccc}
        \hline
                           & \multicolumn{3}{c|}{SMD}                         & \multicolumn{3}{c|}{J-D1}                        & \multicolumn{3}{c|}{J-D2}                        & \multicolumn{3}{c}{SMAP}                         \\ \cline{2-13} 
                           & Precision      & Recall         & F1             & Precision      & Recall         & F1             & Precision      & Recall         & F1             & Precision      & Recall         & F1             \\ \hline 
        DCdetector         & \underline{0.680}    & 0.672          & 0.669          & 0.709          & 0.583          & 0.626          & \textbf{0.956} & 0.897          & 0.923          & 0.594          & 0.613          & 0.597          \\ [0.1cm]
        AnomalyTransformer & 0.439          & \textbf{0.947} & 0.562          & 0.519          & \underline{0.945}    & 0.639          & 0.824          & \underline{0.981}    & 0.891          & 0.610          & 0.947          & 0.699          \\ [0.1cm]
        DVGCRN             & 0.481          & 0.766          & 0.481          & 0.344          & 0.737          & 0.421          & 0.695          & 0.867          & 0.742          & 0.475          & 0.979          & 0.549          \\ [0.1cm]
        OmniAnomaly        & 0.674          & 0.829          & \underline{0.713}    & \textbf{0.957} & 0.868          & \underline{0.899}    & \underline{0.948}    & 0.932          & \underline{0.938}    & 0.789          & 0.984          & 0.819          \\ [0.1cm]
        MSCRED             & 0.444          & 0.562          & 0.407          & 0.880          & 0.806          & 0.819          & 0.927          & 0.944          & 0.932          & \underline{0.838}    & \underline{1.000}    & \underline{0.884}    \\ [0.1cm]
        TranAD             & 0.617          & 0.616          & 0.471          & 0.198          & 0.631          & 0.258          & 0.729          & 0.952          & 0.797          & 0.275          & 0.577          & 0.291          \\ [0.1cm]
        ProS               & 0.153          & 0.785          & 0.214          & 0.505          & 0.731          & 0.534          & 0.796          & 0.861          & 0.805          & 0.412          & 0.973          & 0.468          \\ [0.1cm]
        VAE                & 0.221          & 0.689          & 0.246          & 0.377          & 0.796          & 0.425          & 0.566          & 0.909          & 0.665          & 0.470          & 0.983          & 0.557          \\ \hline
        MACE               & \textbf{0.964} & \underline{0.870}    & \textbf{0.910} & \underline{0.893}    & \textbf{0.984} & \textbf{0.934} & 0.938          & \textbf{0.989} & \textbf{0.961} & \textbf{0.958} & \textbf{1.000} & \textbf{0.977} \\ \hline
        \end{tabular}%
\end{table*}
\textbf{Metrics.} We use three of the most widely-used metrics to evaluate the performance of MACE and baseline methods as many prominent anomaly detection papers \cite{DBLP:conf/icml/ChenTCDDZ22, DBLP:conf/kdd/YangZZW023, DBLP:journals/pvldb/TuliCJ22,ma2021jump} do: the precision, recall and F1 score. The definitions of these metrics are given in Eq.~\ref{Eq:metrics1} - Eq.~\ref{Eq:metrics3}, where $TP$, $FP$ and $FN$ denote true positive, false positive and false negative respectively. 
\begin{gather}
        \operatorname{Precision} = \frac{TP}{TP+FP} \label{Eq:metrics1}\\
        \operatorname{Recall}  =\frac{TP}{FN+TP} \label{Eq:metrics2}\\
        \operatorname{F1}  = 2 * \frac{\operatorname{Precision}*\operatorname{Recall}}{\operatorname{Precision}+\operatorname{Recall}} \label{Eq:metrics3}
\end{gather}
\subsection{Prediction Accuracy}
In this subsection, we conduct extensive experiments to validate that MACE consistently achieves the best F1 score compared to baselines when distinguishing anomalies from different normal patterns with a unified model. Furthermore, in comparison to baselines that tailor a unique model for each subset, MACE demonstrates competitive performance with a unified model. Additionally, owing to the memory-guided pattern extraction method, MACE exhibits commendable performance on previously unseen normal patterns. 

\textbf{Adaptability to multiple normal patterns.} We assume that various subsets in the four datasets contain distinct normal patterns, representing logs for different servers (SMD), services (J-D1 and J-D2), and data for different detector channels (SMAP). During the training stage, every ten subsets in a dataset are grouped together and utilized to train a unified model \textcolor{black}{for both MACE and baselines}. In the testing stage, the trained model is applied to detect anomalies in each corresponding testing subset independently. The average metrics for different subsets are presented in Table \ref{Tab:pq1}, \textcolor{black}{where the best results are highlighted in bold, and the second-best results are underlined}. Since JumpStarter is a signal-based method, multiple normal pattern training is not applicable to it, and thus, JumpStarter is excluded from this analysis. As indicated in Table \ref{Tab:pq1}, despite occasional deviations, MACE achieves the best performance when detecting multiple normal patterns with a unified model. Moreover, MACE consistently achieves the best F1 score compared with all the baselines across the four datasets.
Furthermore, the improvement is substantial: MACE increases the F1 score by an average of 8.7\% compared to the best baseline performance. As illustrated in Fig. \ref{fig:distance}, the subsets in SMD exhibit significant differences from each other, where MACE shows a distinct advantage.
Intuitively, since the normal patterns in J-D2 are very similar to each other, most methods perform well on this dataset and the advantage of MACE is not as obvious as the one on the former dataset. Additionally, MACE attains a high F1 score on SMAP. Considering that most anomalies in SMAP are point anomalies, this result verifies the effectiveness of dualistic convolution in the time domain, which extends the detection capabilities for short-term anomalies and makes them easier to identify. 

Moreover, to verify the unified model can work well on each service, we plot the F1 score for different services when detecting anomalies with a unified model on the Server Machine Dataset in Fig.~\ref{fig:multi}. As shown in Fig.~\ref{fig:multi}, the performance of MACE centered around a pretty well average value, while performances of other methods vary around a broad range across different services. This verifies MACE can capture diverse normal patterns well with a unified model.
 \begin{table*}[]
    \centering
    \vspace{-3mm}
    \renewcommand\arraystretch{1}
    \caption{\label{Tab:pq2}MACE uses a unified model for different patterns, while baselines customize a unique model for each normal pattern.}
    \begin{tabular}{l|ccc|ccc|ccc|ccc}
        \hline
                           & \multicolumn{3}{c|}{SMD}                                                             & \multicolumn{3}{c|}{J-D1}                                                            & \multicolumn{3}{c|}{J-D2}                                                            & \multicolumn{3}{c}{SMAP}                                                            \\ \cline{2-13} 
                           & \multicolumn{1}{c}{Precision} & \multicolumn{1}{c}{Recall} & \multicolumn{1}{c|}{F1} & \multicolumn{1}{c}{Precision} & \multicolumn{1}{c}{Recall} & \multicolumn{1}{c|}{F1} & \multicolumn{1}{c}{Precision} & \multicolumn{1}{c}{Recall} & \multicolumn{1}{c|}{F1} & \multicolumn{1}{c}{Precision} & \multicolumn{1}{c}{Recall} & \multicolumn{1}{c}{F1} \\ \hline
        DCdetector         & 0.836                         & 0.911                      & 0.872                   & 0.766                         & 0.744                      & 0.748                   & \textbf{0.956}                & 0.880                      & 0.913                   & \underline{0.956}                   & 0.989                      & \underline{0.970}            \\ [0.1cm]
        AnomalyTransformer & 0.894                         & \underline{0.955}                & \underline{0.923}             & 0.520                         & 0.918                      & 0.645                   & 0.818                         & \textbf{0.998}             & 0.896                   & 0.941                         & 0.994                      & 0.967                  \\ [0.1cm]
        DVGCRN             & \underline{0.950}                   & 0.883                      & 0.915                   & 0.395                         & 0.806                      & 0.479                   & 0.711                         & 0.852                      & 0.723                   & 0.916                         & 0.920                      & 0.914                  \\ [0.1cm]
        JumpStarter        & 0.904                         & 0.943                      & 0.923                   & \underline{0.921}                   & \underline{0.945}                & \underline{0.933}             & 0.941                         & \underline{0.996}                & \textbf{0.968}          & 0.471                         & 0.995                      & 0.526                  \\ [0.1cm]
        OmniAnomaly        & 0.695                         & 0.877                      & 0.728                   & 0.891                         & 0.940                      & 0.905                   & 0.945                         & 0.974                      & 0.958                   & 0.713                         & 0.963                      & 0.744                  \\ [0.1cm]
        MSCRED             & 0.746                         & 0.744                      & 0.716                   & \textbf{0.975}                & 0.830                      & 0.889                   & \underline{0.949}                   & 0.969                      & 0.958                   & 0.872                         & \underline{1.000}                & 0.923                  \\ [0.1cm]
        TranAD             & 0.926                         & \textbf{0.997}             & \textbf{0.961}          & 0.251                         & 0.918                      & 0.349                   & 0.754                         & 0.965                      & 0.817                   & 0.804                         & \underline{1.000}                & 0.892                  \\ [0.1cm]
        ProS               & 0.146                         & 0.822                      & 0.206                   & 0.422                         & 0.767                      & 0.506                   & 0.763                         & 0.921                      & 0.821                   & 0.447                         & 0.973                      & 0.509                  \\ [0.1cm]
        VAE                & 0.286                         & 0.585                      & 0.255                   & 0.334                         & 0.866                      & 0.385                   & 0.702                         & 0.890                      & 0.763                   & 0.579                         & 0.973                      & 0.648                  \\ \hline
        MACE               & \textbf{0.964}                & 0.870                      & 0.910                   & \underline{0.893}                   & \textbf{0.984}             & \textbf{0.934}          & 0.938                         & 0.989                      & \underline{0.961}             & \textbf{0.958}                & \textbf{1.000}             & \textbf{0.977}         \\ \hline
    \end{tabular}%
    \vspace{-3mm}
\end{table*}

\begin{table*}[]
    \centering
    \renewcommand\arraystretch{1}
    \caption{\label{tab:mc}\textcolor{black}{MACE uses a unified model for different patterns, while baselines customize a unique model for each normal pattern on MC}}
    \begin{tabular}{l|ccccccccc|c}
        \hline
                  & DCdetector             & AnomalyTrans             & DVGCRN & OmniAnomaly & MSCRED & TranAD & ProS           &   VAE   & JumpStarter & MACE \\ \hline
        Precision & \underline{0.984}      & \textbf{1.000}           & 0.125  & 0.681       & 0.841  & 0.784  & 0.681          &  0.583  &   0.473     & 0.908  \\
        Recall    & 0.728                  & 0.870                    & 0.304  & 0.981       & 0.981  & 0.989  & \textbf{1.000} &  0.750  &   0.393     & \underline{0.994}     \\
        F1 Score  & 0.806                  & \underline{0.923}        & 0.147  & 0.782       & 0.878  & 0.864  & 0.772          &  0.639  &   0.393     & \textbf{0.941} \\ \hline
    \end{tabular}%
    \vspace{-3mm}
\end{table*}

\begin{table*}[]
    \centering
    \renewcommand\arraystretch{1}
    \caption{\label{Tab:pq3}The performance of MACE and baselines for unseen normal patterns.}
    \begin{tabular}{l|ccc|ccc|ccc|ccc}
    \hline
                       & \multicolumn{3}{c|}{SMD}                                                             & \multicolumn{3}{c|}{J-D1}                                                            & \multicolumn{3}{c|}{J-D2}                                                            & \multicolumn{3}{c}{SMAP}                                                            \\ \cline{2-13} 
                       & \multicolumn{1}{c}{Precision} & \multicolumn{1}{c}{Recall} & \multicolumn{1}{c|}{F1} & \multicolumn{1}{c}{Precision} & \multicolumn{1}{c}{Recall} & \multicolumn{1}{c|}{F1} & \multicolumn{1}{c}{Precision} & \multicolumn{1}{c}{Recall} & \multicolumn{1}{c|}{F1} & \multicolumn{1}{c}{Precision} & \multicolumn{1}{c}{Recall} & \multicolumn{1}{c}{F1} \\ \hline
    DCdetector         & 0.681                         & 0.685                      & 0.681                   & 0.798                         & 0.771                      & 0.781                   & \underline{0.948}                   & 0.857                      & 0.891                   & 0.700                         & 0.760                      & 0.724                  \\ [0.1cm]
    AnomalyTransformer & 0.490                         & \textbf{0.916}             & 0.622                   & 0.555                         & \textbf{0.948}             & 0.667                   & 0.838                         & \textbf{0.981}             & 0.899                   & 0.586                         & \textbf{1.000}             & 0.678                  \\ [0.1cm]
    DVGCRN             & 0.125                         & 0.798                      & 0.173                   & 0.388                         & \underline{0.894}                & 0.478                   & 0.619                         & 0.893                      & 0.664                   & 0.444                         & \textbf{1.000}             & 0.525                  \\ [0.1cm]
    OmniAnomaly        & \underline{0.686}                   & 0.780                      & \underline{0.701}             & \textbf{0.976}                & 0.824                      & \underline{0.880}             & 0.932                         & 0.953                      & \underline{0.941}             & 0.735                         & 0.986                      & 0.794                  \\ [0.1cm]
    MSCRED             & 0.418                         & 0.593                      & 0.409                   & 0.828                         & 0.818                      & 0.806                   & 0.931                         & 0.952                      & 0.939                   & \underline{0.839}                   & \textbf{1.000}             & \underline{0.896}            \\ [0.1cm]
    TranAD             & 0.255                         & 0.643                      & 0.265                   & 0.127                         & 0.546                      & 0.198                   & 0.516                         & 0.659                      & 0.546                   & 0.205                         & \textbf{1.000}             & 0.302                  \\ [0.1cm]
    ProS               & 0.154                         & 0.808                      & 0.215                   & 0.475                         & 0.770                      & 0.564                   & 0.789                         & 0.933                      & 0.855                   & 0.412                         & 0.979                      & 0.469                  \\ [0.1cm]
    VAE                & 0.193                         & 0.789                      & 0.270                   & 0.339                         & 0.875                      & 0.386                   & 0.661                         & 0.884                      & 0.721                   & 0.433                         & 0.979                      & 0.500                  \\ \hline
    MACE               & \textbf{0.915}                & \underline{0.835}                & \textbf{0.863}          & \underline{0.972}                   & 0.829                      & \textbf{0.885}          & \textbf{0.963}                & \underline{0.968}                & \textbf{0.964}          & \textbf{0.954}                & \underline{0.996}                & \textbf{0.973}         \\ \hline
    \end{tabular}%
    \vspace{-3mm}
\end{table*}

\textbf{Competitive performance compared to customizing a unique model.} In this experiment, MACE employs a unified model for every ten different normal patterns, while baselines customize a unique model for each subset. As depicted in Table \ref{Tab:pq2} and Table \ref{tab:mc}, MACE achieves comparable performance with the strongest state-of-the-art methods. It is noteworthy that MACE captures ten different patterns simultaneously with a single model, a factor that generally hinders model performance \cite{park2020learning}, while baselines customize a unique model for each normal pattern. The negative impact of multiple normal patterns is further evident when comparing Table \ref{Tab:pq1} and Table \ref{Tab:pq2}: when normal patterns are diverse (e.g., in SMD), the baselines in Table \ref{Tab:pq2}, where they tailor a unique model for each normal pattern, exhibit big strength compared to their performance in Table \ref{Tab:pq1}, where they learn a unified model for multiple normal patterns. In contrast, when normal patterns are similar (e.g., in J-D2), the baseline performance gaps between Tables \ref{Tab:pq1} and \ref{Tab:pq2} are narrow. From this comparison, it can be concluded that the diversity of normal patterns hinders model performance when using a unified model for multiple patterns.
Thus, it is tolerable for MACE to exhibit a somewhat lower F1 score on SMD, considering that the normal patterns of SMD are the most diverse among the four datasets.

\begin{figure*}[t]
    \centering 
    \vspace{-3mm}
    \subfigure[Time (s) and memory overhead (*10kB).]{
    \includegraphics[width=0.3\linewidth]{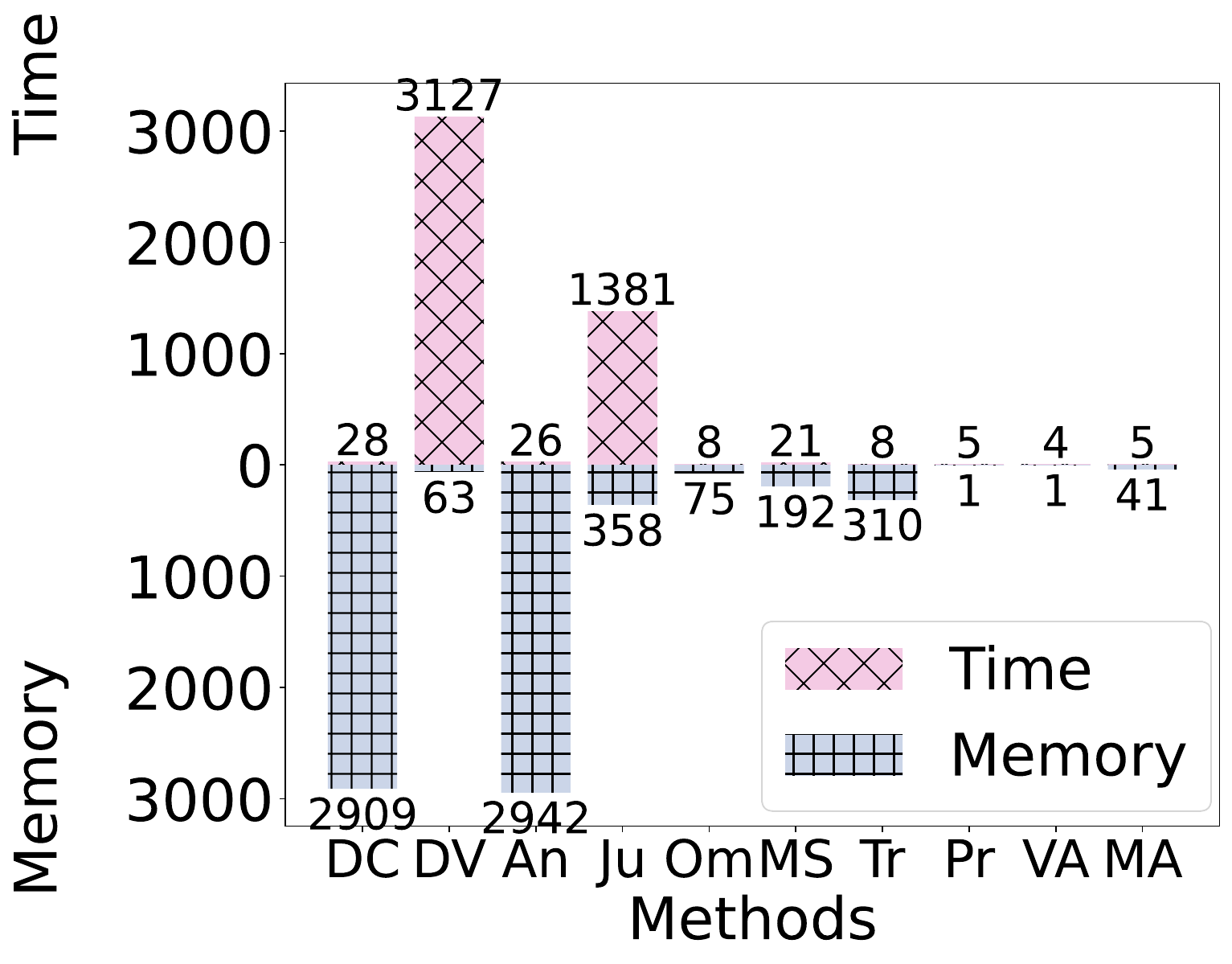}
    \label{fig:overhead}
    }
    \hfill
    \subfigure[The impact of $\gamma_t$ and $\gamma_f$ on MACE performance.]{
    \includegraphics[width=0.3\linewidth]{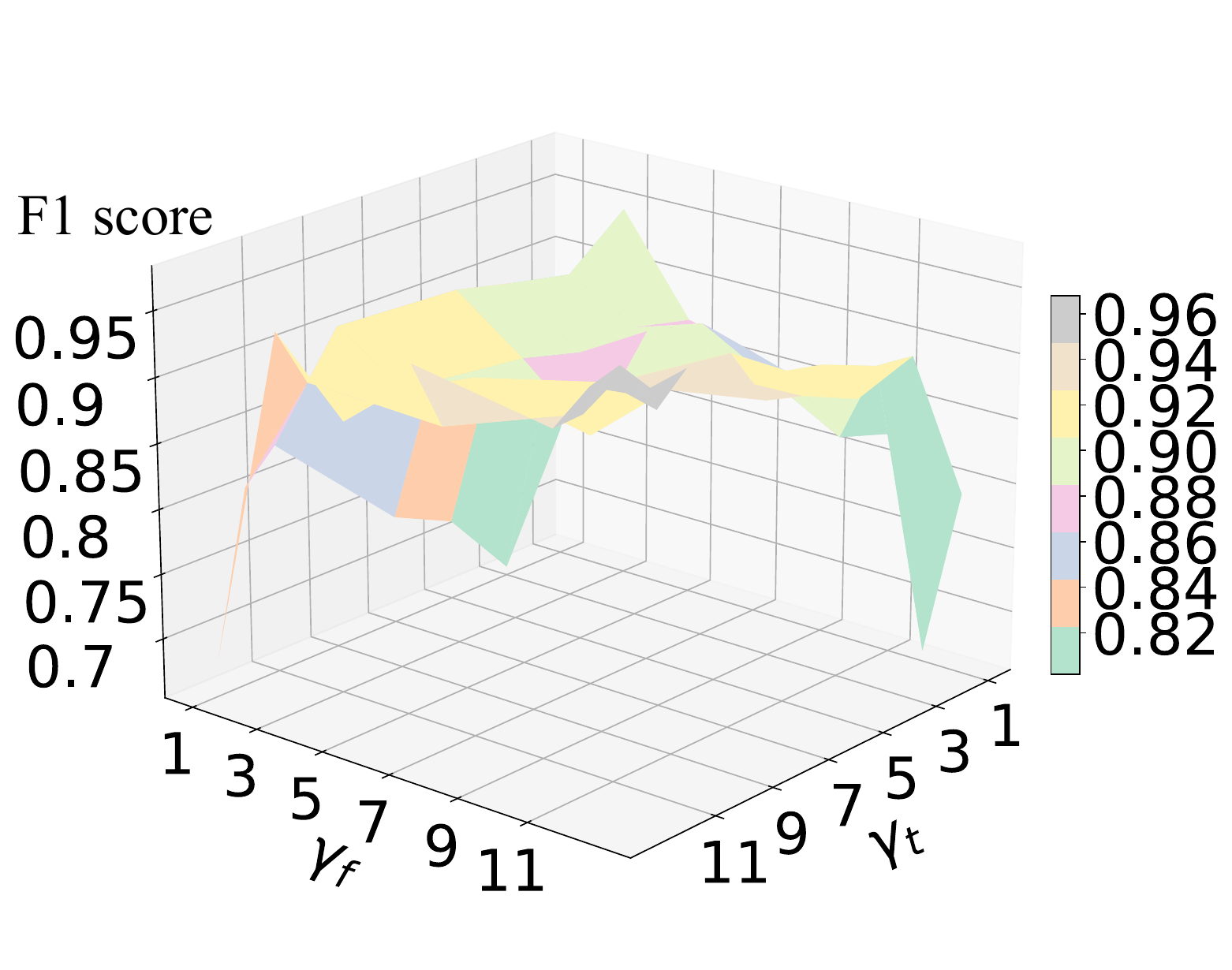}
    \label{fig:tgg}
    }
    \hfill
    \subfigure[The impact of $\gamma_t$ and $\sigma_t$ on MACE performance.]{
    \includegraphics[width=0.3\linewidth]{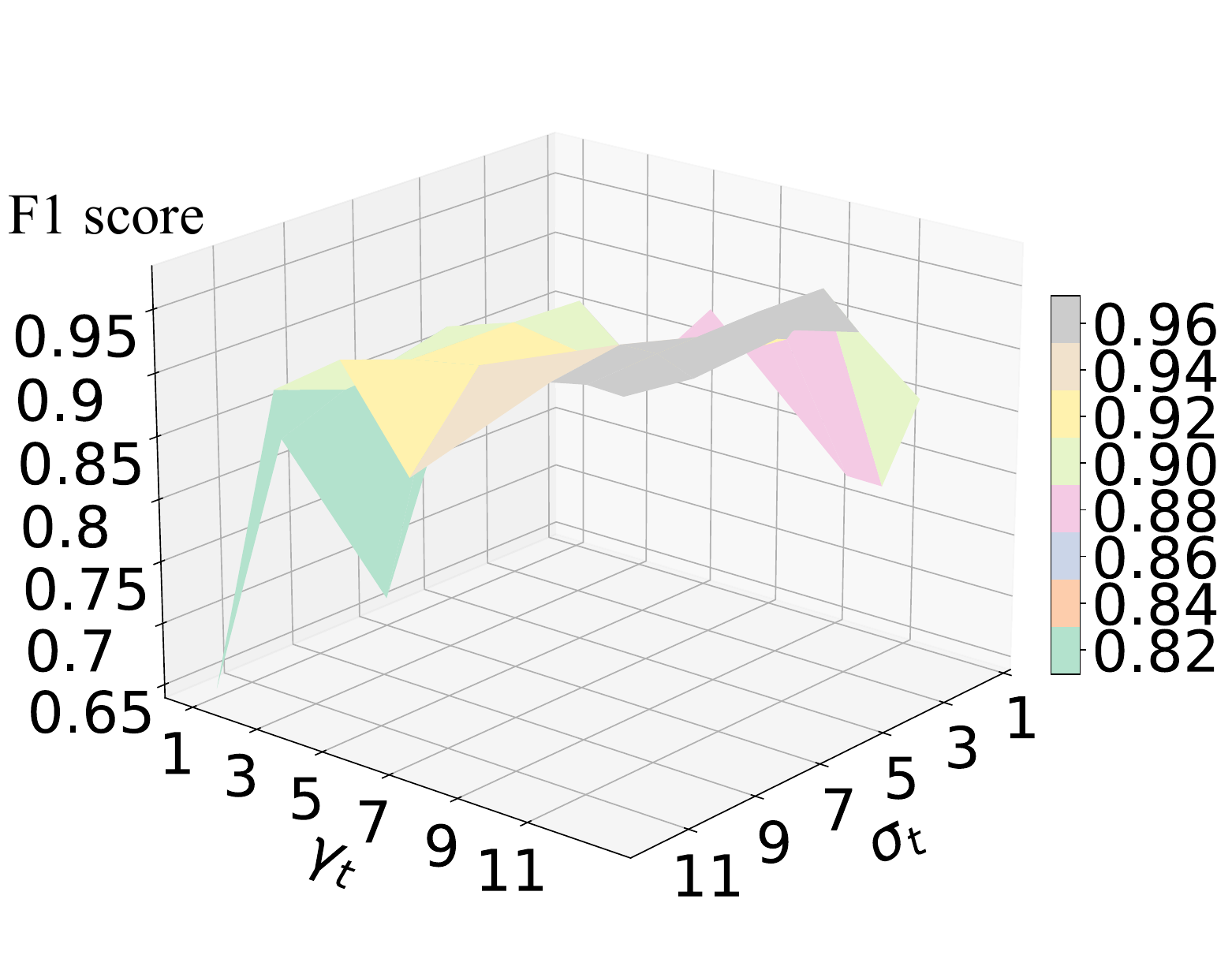}
    \label{fig:tgs}
    }
    \hfill
    \subfigure[The impact of $\gamma_f$ and $\sigma_f$ on MACE performance.]{
    \includegraphics[width=0.3\linewidth]{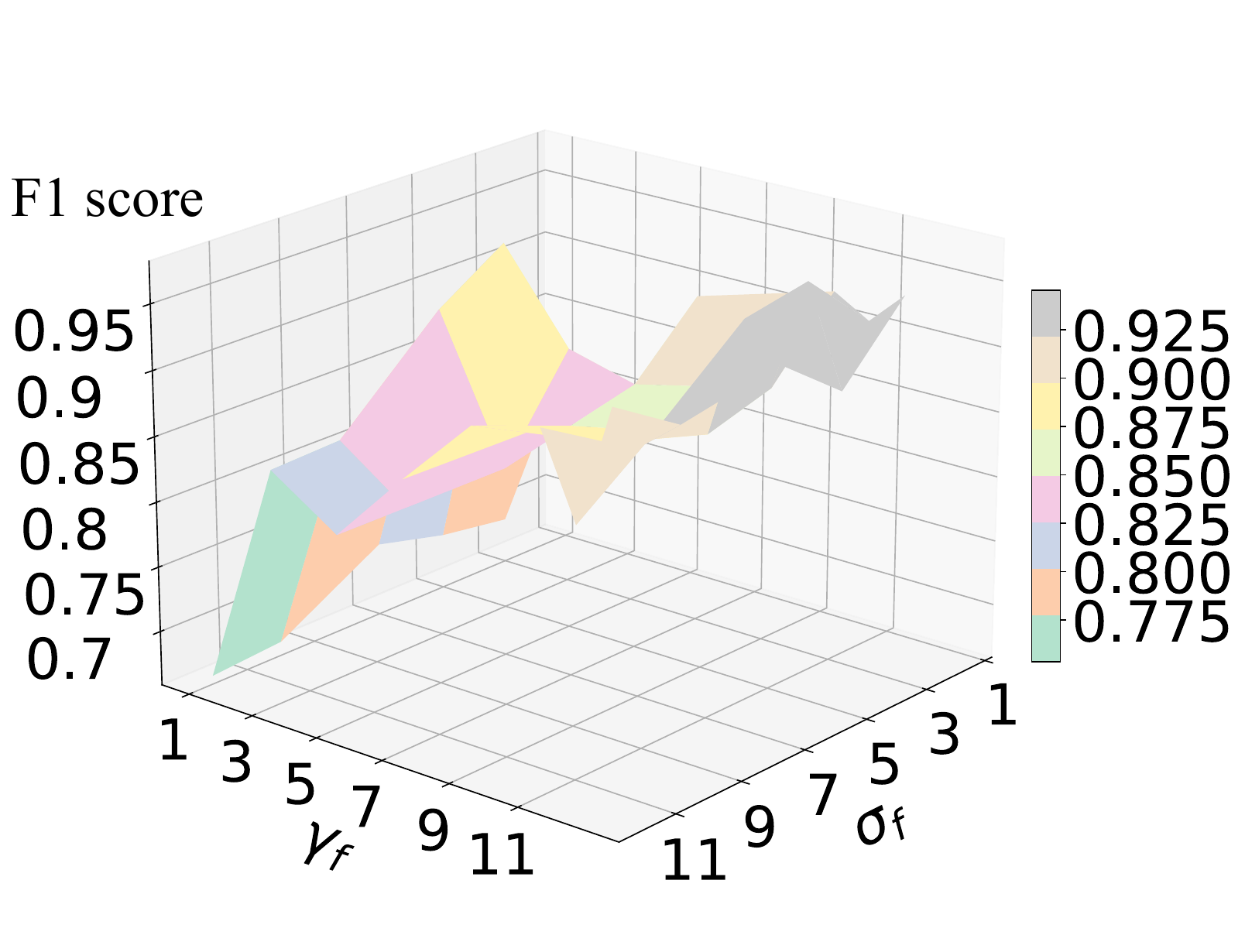}
    \label{fig:fgs}
    }
    \hfill
    \subfigure[The impact of kernel size and $\gamma_t$ on MACE performance.]{
    \includegraphics[width=0.3\linewidth]{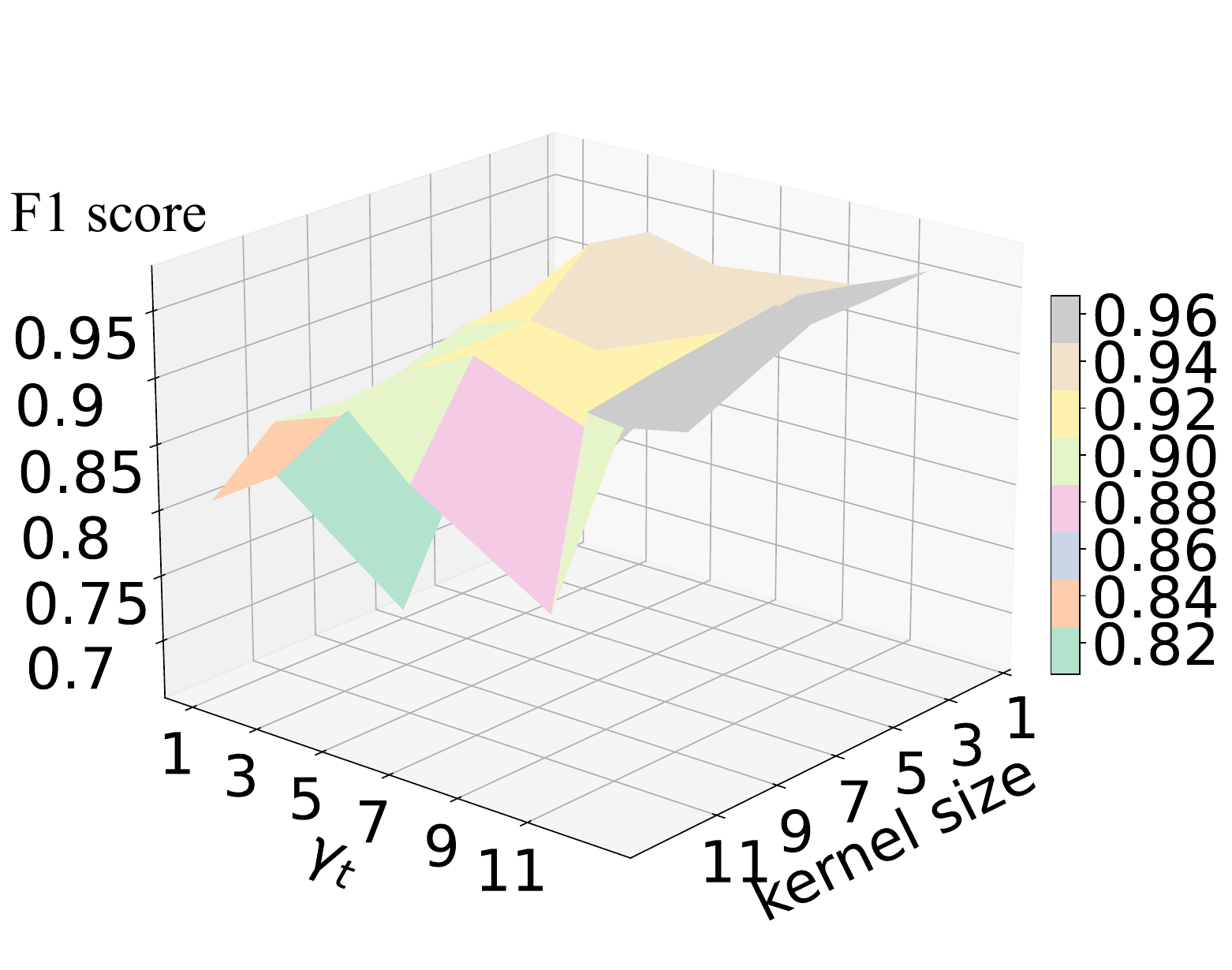}
    \label{fig:tkg}
    }
    \hfill
    \subfigure[The impact of number of basis in a subset and $\gamma_f$ on MACE performance.]{
    \includegraphics[width=0.3\linewidth]{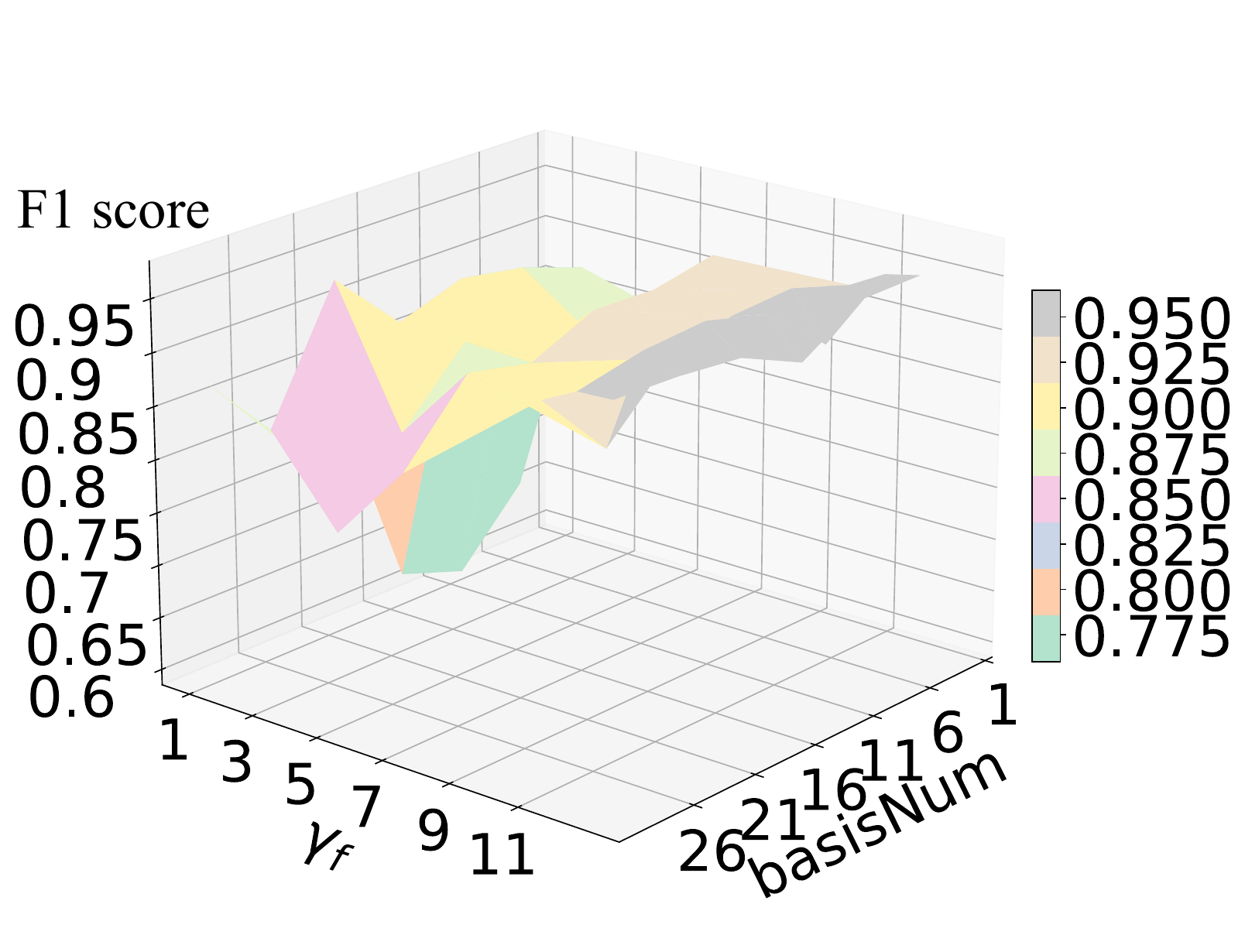}
    \label{fig:tbg}
    }
    \hfill
    
    \caption{Due to the space limitation, we use the first two letters in method names as a shorthand. $\gamma_t$ and $\gamma_f$ represent the powers in dualistic convolution for the time domain and frequency domain respectively. $\sigma_t$ and $\sigma_f$ represent the scaling factors in dualistic convolution for the time domain and frequency domain respectively.  (a) The time and memory overhead of MACE and baselines. (b) The F1 score of MACE for grid search of $\gamma_t$ and $\gamma_f$. (c) The F1 score of MACE for grid search of $\gamma_t$ and $\sigma_t$. (d) The F1 score of MACE for grid search of $\gamma_f$ and $\sigma_f$. (e) The F1 score of MACE for grid search of dualistic convolution kernel size in the time domain and $\gamma_t$. (f) The F1 score of MACE for grid search of the number of bases in a subset and $\gamma_f$. }
    \vspace{-2em}
\end{figure*}

\textbf{MACE performance on unseen normal patterns.} As mentioned earlier, different subsets are assumed to represent different normal patterns, and every ten subsets in a dataset are divided into a group. MACE and all the baselines are trained on one group and tested on another. The results are presented in Table \ref{Tab:pq3}, where the best performances are bolded, and the second-best performances are underlined. Since JumpStarter is a signal-based method, training on one group while testing on another is not applicable to it, and thus, it is not included in the table. As shown in Table \ref{Tab:pq3}, MACE consistently achieves the highest F1 score on the four datasets. The performance of MACE when the normal patterns are diverse, such as in SMD, is lower than when the normal patterns are similar, such as in J-D2. When the distance between different normal patterns is small, MACE can achieve similar F1 scores to those when MACE is trained and tested on the same group (i.e., the performance on J-D2 and SMAP).

\begin{table*}[htbp]
    \centering
    \renewcommand\arraystretch{1}
    \caption{\label{Tab:abl}The performances of MACE when removing different modules}
    \begin{tabular}{l|ccc|ccc|ccc|ccc}
        \hline
        \multicolumn{1}{c|}{\multirow{2}{*}{Remove module}} & \multicolumn{3}{c|}{SMD}                                                             & \multicolumn{3}{c|}{J-D1}                                                            & \multicolumn{3}{c|}{J-D2}                                                            & \multicolumn{3}{c}{SMAP}                                                            \\ \cline{2-13} 
        \multicolumn{1}{c|}{}                               & \multicolumn{1}{c}{Precision} & \multicolumn{1}{c}{Recall} & \multicolumn{1}{c|}{F1} & \multicolumn{1}{c}{Precision} & \multicolumn{1}{c}{Recall} & \multicolumn{1}{c|}{F1} & \multicolumn{1}{c}{Precision} & \multicolumn{1}{c}{Recall} & \multicolumn{1}{c|}{F1} & \multicolumn{1}{c}{Precision} & \multicolumn{1}{c}{Recall} & \multicolumn{1}{c}{F1} \\ \hline
        Context-aware DFT \& IDFT                            & 0.762                         & 0.813                      & 0.762                   & 0.624                         & 0.887                      & 0.689                   & \underline{0.958}                   & 0.953                      & 0.953                   & 0.775                         & \textbf{1.000}             & 0.831                  \\ [0.1cm] 
        Dualistic Convolution (F)                           & 0.187                         & \textbf{0.943}             & 0.184                   & \underline{0.855}                   & 0.854                      & 0.820                   & 0.857                         & 0.933                      & 0.886                   & 0.681                         & 0.990                      & 0.713                  \\ [0.1cm]
        Dualistic Convolution (T)                           & 0.046                         & 0.842                      & 0.084                   & 0.089                         & \underline{0.946}                & 0.152                   & 0.208                         & 0.440                      & 0.250                   & 0.682                         & 0.989                      & 0.720                  \\ [0.1cm]
        Frequency Characterization                          & \underline{0.894}                   & 0.860                      & \underline{0.868}             & 0.838                         & 0.940                      & \underline{0.857}             & \textbf{0.970}                & \underline{0.980}                & \textbf{0.975}          & \underline{0.944}                   & 1.000                      & \underline{0.967}            \\ [0.1cm] 
        Pattern extraction                              & 0.714                         & 0.702                      & 0.696                   & 0.667                         & 0.914                      & 0.740                   & 0.957                         & 0.953                      & 0.954                   & 0.770                         & 0.970                      & 0.797                  \\ [0.1cm] \hline 
        MACE                                                & \textbf{0.964}                & \underline{0.870}                & \textbf{0.910}          & \textbf{0.893}                & \textbf{0.984}             & \textbf{0.934}          & 0.938                         & \textbf{0.989}             & \underline{0.961}             & \textbf{0.958}                & \textbf{1.000}             & \textbf{0.977}         \\ \hline 
    \end{tabular}%
    \vspace{-2mm}
\end{table*}

\subsection{Efficiency Analysis}
We evaluated both time and memory overhead on a server equipped with a configuration comprising 32 Intel(R) Xeon(R) CPU E5-2620 @ 2.10GHz CPUs and 2 K80 GPUs.
For neural network-based methods, we employed a profiling tool to assess their memory overhead. In the case of JumpStarter, a signal-based method, we record its maximum memory consumption during the inference process. The time overhead was calculated based on the training time of each method on a subset group of the SMD dataset.
The results, depicted in Fig. \ref{fig:overhead}, reveal that MACE's time overhead is competitive with some very simple methods, such as VAE and ProS based on VAE, while MACE's F1 scores significantly surpass the ones of them across all four datasets. Regarding memory overhead, MACE's value is higher than that of a two-layer VAE and ProS based on a two-layer VAE. However, MACE's memory overhead is considerably lower than that of other deep neural networks.
These findings underscore MACE's efficiency in terms of both time and memory usage, positioning it favorably among deep neural-network-based methods and showcasing superior performance in terms of anomaly detection as evidenced by its higher F1 scores across diverse datasets.

\subsection{Ablation Study}
\label{sec:abl}
We conducted experiments to assess the effectiveness of individual modules within MACE by removing them individually. When the context-aware Discrete Fourier Transform (DFT) and Inverse DFT (IDFT) modules were removed, they were replaced with conventional DFT and IDFT.  \textcolor{black}{When making ablation experiments for other modules, we compare the completed MACE with the one that has removed them.}
The results are displayed in Table \ref{Tab:abl}, where "Dualistic Convolution (F)" and "Dualistic Convolution (T)" correspond to dualistic convolution in the frequency and time domains, respectively.
As depicted in Table \ref{Tab:abl}, the complete MACE model exhibits considerable superiority over its variants. It is worth noting that when context-aware DFT and IDFT are substituted with vanilla counterparts, MACE's performance takes a sharp nosedive. The computational and memory overhead of vanilla DFT and IDFT increases because they introduce more Fourier bases, yet the performance deteriorates. This observation aligns with our earlier theoretical analysis.
Moreover, this experiment underscores the effectiveness of the diverse normal pattern adaptability facilitated by the pattern extraction mechanism. This module significantly enhances performance on SMD, characterized by diverse normal patterns, while showing marginal improvement on J-D2, where the normal patterns are similar. Similarly, the frequency characterization module makes a substantial contribution on SMD but is of limited use on J-D2, given its multi-pattern extraction nature.
In summary, the module-by-module evaluation reaffirms the crucial role played by various components in MACE and their impact on anomaly detection performance across different datasets.

\subsection{Hyperparameter Study}
We employed grid search to investigate the influence of critical hyperparameters on the performance of MACE. Fig.\ref{fig:tgg}-Fig.\ref{fig:tbg} present the F1 scores corresponding to different combinations of pairwise hyperparameters. The search ranges for $\gamma$ in the time domain and frequency domain, $\sigma$ in the time domain and frequency domain, kernel size, and the number of Fourier bases in Context-aware DFT and IDFT were set to \{1, 3, 5, 7, 11, 12, 13\}, \{3, 5, 7, 10, 12\}, \{3, 5, 7, 11, 13\}, and \{5, 10, 15, 20, 25, 30\} respectively. 

\textbf{Impact of $\gamma_t$ and $\gamma_f$}: When $\gamma_t$ and $\gamma_f$ are set to 1, dualistic convolution degenerates into a standard convolution, essentially nullifying its contribution. Consequently, MACE's performances with $\gamma_t$ and $\gamma_f$ set to 1 are unsatisfied. In general, the performance of MACE improves as $\gamma_t$ and $\gamma_f$ increase, as depicted in Fig.\ref{fig:tgs} and Fig.\ref{fig:fgs}. However, it's important to note that $\gamma$ cannot grow infinitely, as excessively large $\gamma$ values can lead to gradient explosions. Thus, setting $\gamma$ within the search space mentioned above is a safe approach.

\textbf{Impact of $\sigma_t$ and $\sigma_f$}: These scaling factors are introduced to mitigate gradient explosions. As demonstrated in Fig.\ref{fig:tgs}-Fig.\ref{fig:fgs}, MACE's performance remains stable across various values of $\sigma$.

\textbf{Impact of dualistic convolution kernel size in the time domain}: Intuitively, the performance of MACE initially improves and then declines as the kernel size increases, as shown in Figure \ref{fig:tkg}. That is because when the kernel size increases from a small value, it makes anomalies more prominent and easier to detect. However, when the kernel size becomes excessively large, the dualistic convolution in the time domain distorts the original time series and detrimentally affects model performance.

\textbf{Impact of the number of Fourier bases in context-aware DFT and IDFT}: The performance of MACE generally follows an increasing-then-decreasing pattern as the number of bases grows. As analyzed theoretically in Section.\ref{sec:PaExtr}, when the number of bases increases from a small value, both the reconstruction of normal patterns and anomalies improve, but the enhancement in normality reconstruction is more pronounced. However, when the number of bases becomes relatively large, the improvement in normality reconstruction becomes marginal, while the effect on anomaly reconstruction becomes significant. Therefore, the performance initially increases and then declines with the growth of the number of bases.


\section{Conclusion}
In this work, we address the challenges of detecting anomalies from diverse normal patterns with a unified and efficient model as well as improve the short-term anomaly sensitivity by proposing MACE. MACE exhibits three key characteristics: (i) a pattern extraction mechanism allowing the model to detect anomalies by the correlation between the data sample and its service normal pattern and adapt to the diverse nature of normal patterns; (ii) a dualistic convolution mechanism that amplifies anomalies in the time domain and hinders the reconstruction in the frequency domain; (iii) the utilization of the inherent sparsity and parallelism of the frequency domain to enhance model efficiency.
We substantiate our approach through both mathematical analysis and extensive experiments, demonstrating that selecting a subset of Fourier bases based on normal patterns yields superior performance compared to utilizing the complete spectrum. Comprehensive experiments confirm MACE's proficiency in effectively handling diverse normal patterns, showcasing optimal performance with high efficiency when benchmarked against state-of-the-art methods.
\section*{Acknowledgment}
This work was supported by the Key Research Project of Zhejiang Province under Grant 2022C01145, the National Science Foundation of China under Grants 62125206 and U20A20173, the Guangzhou-HKUST(GZ) Joint Funding Program under Grant
 2024A03J0620, and in part by Alibaba Group through Alibaba Research Intern Program.

\balance
\bibliography{IEEEexample}
\bibliographystyle{IEEEtran}

\end{document}